\documentclass[10pt,twocolumn,letterpaper]{article}

\usepackage[pagenumbers]{cvpr} % To force page numbers, e.g. for an arXiv version

\usepackage{graphicx}
\usepackage{amsmath}
\usepackage{amssymb}
\usepackage{booktabs}
\usepackage{pifont}
\usepackage[dvipsnames]{xcolor}
\usepackage[accsupp]{axessibility}

\usepackage[pagebackref,breaklinks,colorlinks]{hyperref}

\usepackage[capitalize]{cleveref}
\crefname{section}{Sec.}{Secs.}
\Crefname{section}{Section}{Sections}
\Crefname{table}{Table}{Tables}
\crefname{table}{Tab.}{Tabs.}

\newcommand{\customfootnotetext}[2]{{% Group to localize change to footnote
  \renewcommand{\thefootnote}{#1}% Update footnote counter representation
  \footnotetext[0]{#2}}}% Print footnote text

\def\cvprPaperID{3647} % *** Enter the CVPR Paper ID here
\def\confName{CVPR}
\def\confYear{2023}

\begin{document}
\title{Frequency-Modulated Point Cloud Rendering with Easy Editing}

\author{
Yi Zhang\textsuperscript{1*} \hspace{2mm}
Xiaoyang Huang\textsuperscript{1*} \hspace{2mm}
Bingbing Ni\textsuperscript{1$\dagger$} \hspace{2mm} 
Teng Li\textsuperscript{2} \hspace{2mm} 
Wenjun Zhang\textsuperscript{1} \\
\textsuperscript{1}Shanghai Jiao Tong University, Shanghai 200240, China \qquad \textsuperscript{2}Anhui University \\
{\tt\small \{yizhangphd, huangxiaoyang, nibingbing\}@sjtu.edu.cn}
}

% \maketitle
\twocolumn[{%
\renewcommand\twocolumn[1][]{#1}%
\maketitle
\begin{center}
    \centering
    \vspace{-0.1in}
    \captionsetup{type=figure}
    \includegraphics[width=\linewidth]{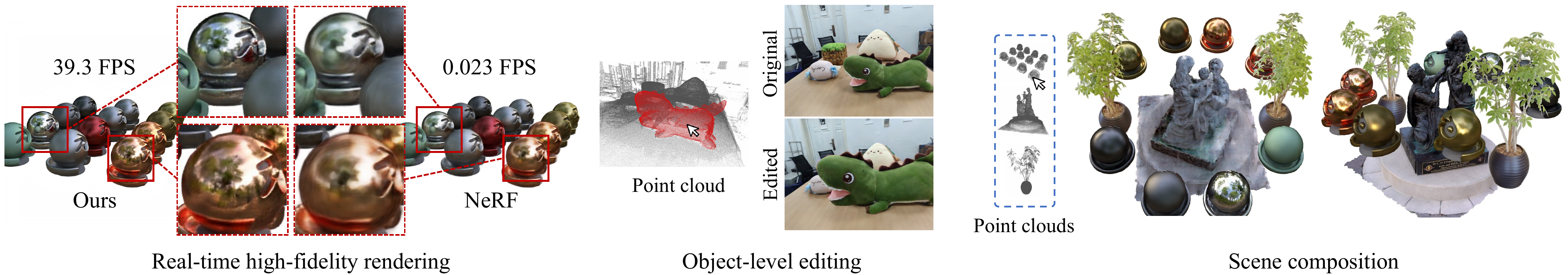}
    \vspace{-0.2in}
    \captionof{figure}{We propose a novel frequency-modulated point cloud rendering pipeline, which enables high fidelity local detail reconstruction, real-time rendering and user-friendly editing. 
    Specifically, we can render sharper textures with 1700$\times$ faster than NeRF~\cite{mildenhall2021nerf} and support both object-level editing and scene composition.}
    \label{fig:cover}
\end{center}%
}]

\customfootnotetext{*}{Equal contribution.}
\customfootnotetext{${\dagger}$}{Corresponding author: Bingbing Ni.}
%%%%%%%%% ABSTRACT
\begin{abstract} 
We develop an effective point cloud rendering pipeline 
% CR
% trained with 2D image supervision,
for novel view synthesis,
which enables high fidelity local detail reconstruction, real-time rendering and user-friendly editing.
In the heart of our pipeline is an adaptive frequency modulation module called \emph{Adaptive Frequency Net (AFNet)}, which utilizes a hypernetwork to learn the local texture frequency encoding that is consecutively injected into adaptive frequency activation layers to modulate the implicit radiance signal.
This mechanism improves the frequency expressive ability of the network with richer frequency basis support, only at a small computational budget.
To further boost performance, a preprocessing module is also proposed for point cloud geometry optimization via point opacity estimation.
In contrast to implicit rendering, our pipeline supports high-fidelity interactive editing based on point cloud manipulation.
Extensive experimental results on NeRF-Synthetic, ScanNet, DTU and Tanks and Temples datasets demonstrate the superior performances achieved by our method in terms of PSNR, SSIM and LPIPS, in comparison to the state-of-the-art.
Code is released at \href{https://github.com/yizhangphd/FreqPCR}{https://github.com/yizhangphd/FreqPCR}.
\end{abstract}

\vspace{-0.15in}
%%%%%%%%% BODY TEXT
\section{Introduction}
\label{sec:intro}
Photo-realistic rendering and editing of 3D representations is a key problem in 3D computer vision and graphics with numerous applications, such as computer games, VR/AR, and video creation.
In particular, recently introduced neural radiance field (NeRF)~\cite{mildenhall2021nerf} has inspired some follow-up works aiming to editable rendering~\cite{yang2021learning, kobayashi2022decomposing,tang2022compressible,liu2021editing,wang2022clip,kania2022conerf}.
Due to the deeply coupled black-box network, NeRF-based object-level editing usually requires a pre-trained segmentation model to separate the objects to be edited~\cite{yang2021learning, kobayashi2022decomposing}.
Although some recent voxel-based variants of NeRF~\cite{tang2022compressible, yu2021plenoctrees} achieve multi-scene composition, they still lack the ability to extract target objects from voxels.
% \vspace{-0.03in}

In contrast to implicit rendering, point cloud rendering~\cite{dai2020neural, oursaaai, aliev2020neural, rakhimov2022npbg++, yifan2019differentiable, ruckert2022adop,li2022read} is a promising editable rendering model.
On the one hand, explicit 3D representations are better for interactive editing.
On the other hand, the geometric priors provided by point clouds can help us avoid massive sampling in volume rendering methods, which can meet the requirements of some real-time applications.
As a class of representative point cloud rendering methods, NPBG and NPBG++~\cite{aliev2020neural, rakhimov2022npbg++} achieve real-time rendering by using point-wise features for encoding appearance information and an U-Net~\cite{ronneberger2015u} for decoding, respectively.
However, the parameter quantity increases with the size of point cloud, which may limit their application due to the excessive computational and memory complexity.
Huang \textit{et~al.}~\cite{oursaaai} combine point clouds with implicit rendering, where  explicit point clouds are used to estimate the geometry, and implicit radiance mapping is used to predict view-dependent appearance of surfaces.
However, quantitative evaluation of its rendering results is significantly lower than that of implicit rendering methods such as NeRF~\cite{mildenhall2021nerf}, mainly due to the following reasons: 1)
the color of each viewing ray only depends on a single surface point, thus without multiple sample color aggregation for error attenuation, surface based rendering techniques require radiance mapping to have a more precise and expressive frequency encoding ability; and
2) defects of the point cloud geometry reconstructed by MVSNet~\cite{yao2018mvsnet} cause wrong surface estimation.
To this end, we introduce Adaptive Frequency Net (AFNet) to improve frequency expression ability of the radiance mapping and a preprocessing module for point cloud geometry optimization.

Radiance mapping, also known as radiance field, is a type of Implicit Neural Representation (INR). There have been some studies~\cite{fathony2020multiplicative, tancik2020fourier, sitzmann2020implicit, yuce2022structured, rahaman2019spectral,benbarka2022seeing,tancik2021learned} on the expressive power and inductive bias of INRs.
The standard Multi-layer Perceptrons (MLPs) with ReLU activation function are well known for the strong spectral bias towards reconstructing low frequency signals~\cite{rahaman2019spectral}.
Some recent works~\cite{tancik2020fourier, sitzmann2020implicit, fathony2020multiplicative} introduce strategies to enhance the high-frequency representation of MLPs from a global perspective.
However, from a local perspective, the frequencies of a 3D scene are region-dependent and most real objects are composed by both weak textured regions and strong textured ones.
Motivated by this, we design a novel adaptive frequency modulation mechanism based on HyperNetwork architecture~\cite{ha2016hypernetworks}, which learns the local texture frequency and injects it into adaptive frequency activation layers to modulate the implicit radiance signal.
The proposed mechanism can predict suitable frequency without frequency supervision and modulate the radiance signal with adaptive frequency basis support to express more complex textures at negligible computational overhead.

Previous surface point-based works~\cite{aliev2020neural, rakhimov2022npbg++, oursaaai} could not optimize the point cloud geometry because they keep only the closest point as a surface estimation for each ray.
But if we sample multiple points per ray during rendering, it will greatly reduce the rendering speed.
Therefore, we use the volume rendering method as a preprocessing module to optimize the point cloud geometry.
Specifically, we keep more points in the pixel buffer and learn the opacity of each point based on volume rendering.
We find in our experiments that for some poorly reconstructed scenes, point cloud geometry optimization can improve the rendering PSNR by 2-4dB and avoid rendering artifacts.

For rigid object editing, we follow the deformation field construction~\cite{tretschk2021non, park2021nerfies,pumarola2021d,park2021hypernerf,peng2021animatable} to render the edited point cloud. Point cloud can be seen as a bridge between user editing and deformation field to achieve interactive editing and rendering. Users only need to edit the point cloud, and the deformation field between the original 3D space and the deformed space is easy to obtain by querying the corresponding transformations performed on point cloud.
Moreover, to avoid cross-scene training in multi-scene composition application, we develop a masking strategy based on depth buffer to combine multiple scenes in pixel level.

We evaluate our method on NeRF-Synthetic~\cite{mildenhall2021nerf}, ScanNet~\cite{dai2017scannet}, DTU~\cite{jensen2014large} and Tanks and Temples~\cite{knapitsch2017tanks} datasets and comprehensively compare the proposed method with other works in terms of performance (including PSNR, SSIM and LPIPS), model complexity, rendering speed, and editing ability.
Our performance outperforms NeRF~\cite{mildenhall2021nerf} and all surface point-based  rendering methods\cite{aliev2020neural, rakhimov2022npbg++, oursaaai}, and is comparable to Compressible-composable NeRF (CCNeRF), \emph{i.e.}, the latest NeRF-based editable variant.
We achieve a real-time rendering speed of 39.27 FPS on NeRF-Synthetic, which is 1700$\times$ faster than NeRF and 37$\times$ faster than CCNeRF.
We also reproduce the scene editing of Object-NeRF~\cite{yang2021learning} and CCNeRF~\cite{tang2022compressible} on ToyDesk~\cite{yang2021learning} and NeRF-Synthetic~\cite{mildenhall2021nerf} dataset, respectively.
As shown in Fig.~\ref{fig:cover}, we achieve real-time rendering with sharper details and user-friendly editing.
The above results demonstrate that our method is comprehensive in terms of both rendering and editing and has great application potentials.

% \vspace{-0.1in}
\section{Related Work}
% \vspace{-0.1in}
\label{sec:relat}
\paragraph{Implicit Neural Representations.}
The goal of INRs is to encode a continuous target signal using a neural network, by representing the mapping between input coordinates and signal values.
This technique has been used to represent various objects such as images~\cite{stanley2007compositional, nguyen2015deep}, shapes~\cite{park2019deepsdf, genova2019learning, chen2019learning}, scenes~\cite{mildenhall2021nerf, sitzmann2019scene}, and textures~\cite{oechsle2019texture, henzler2020learning}.
Tancik \textit{et~al.}~\cite{tancik2020fourier} propose Fourier Feature Network (FFN), showing that feeding input coordinates through a Fourier feature mapping enables MLP to learn high-frequency functions in low dimensional domains.
Sitzmann \textit{et~al.}~\cite{sitzmann2020implicit} propose to use MLP with sinusoidal activations.
Fathony \textit{et~al.} propose MFN~\cite{fathony2020multiplicative}, which employs a Hadamard product between linear layers and nonlinear activation functions.
In \cite{yuce2022structured}, Y{\"u}ce \textit{et~al.} prove that FFN~\cite{tancik2020fourier}, SIREN~\cite{sitzmann2020implicit} and MFN~\cite{fathony2020multiplicative} have the same expressive power as a structured signal dictionary.
% whose atoms are sinusoids with frequencies equal to sums and differences of the integer harmonics of the mapping frequencies.
% CR-delete
% Different from previous work, we design an adaptive frequency modulation mechanism to predict suitable frequency and adaptively modulate the radiance signal.

\vspace{-0.15in}
\paragraph{3D Representations for View Synthesis.}
Implicit representations often use neural networks to encode 3D scenes~\cite{zhang2021learning,wang2021neus,oechsle2021unisurf,mildenhall2021nerf,park2019deepsdf,niemeyer2020differentiable,wang2021ibrnet, shi2022garf}.
NeRF~\cite{mildenhall2021nerf} uses a 5D function to represent the scene and applies volumetric rendering for view synthesis.
Many recent works combine NeRF with some explicit representation to improve rendering quality or speed.
NSVF~\cite{liu2020neural} uses a sparse voxel octree to achieve fast rendering.
Yu \textit{et~al.} propose PlenOctrees~\cite{yu2021plenoctrees}, a data structure derived from NeRF\cite{mildenhall2021nerf} which enables highly efficient rendering.
Plenoxels\cite{fridovich2022plenoxels}  represents a scene as a sparse 3D grid with spherical harmonics.
Xu~\textit{et~al.} \cite{xu2022point} propose Point-NeRF, which is a localized neural representation, combining volumetric radiance fields with point clouds.
Instant-NGP~\cite{muller2022instant} uses multi-resolution hashing for efficient encoding and also leads to high compactness.
TensoRF~\cite{chen2022tensorf} factorizes the scene tensor into lower-rank components.
Huang \textit{et al.}~\cite{oursaaai} combine point cloud with implicit radiance mapping to achieve the state-of-the-art performance of point cloud rendering. However, their performance is still worse than NeRF~\cite{mildenhall2021nerf}. In this work, we greatly improve the performance of point cloud rendering by introducing AFNet and a preprocessing module for geometry optimization.

\vspace{-0.1in}
\paragraph{Editable Rendering.} 
Editing ability is important for 3D representations.
Some recent works~\cite{guo2020object, yang2021learning, yuan2022nerf, yang2022neumesh,sun2022fenerf,xu2022deforming} improve NeRF to editable version.
% Guo \textit{et~al.}\cite{guo2020object} propose a bottom-up method by learning one scattering field per-object and enables rendering scenes with moving objects and lights.
% NSVF\cite{liu2020neural} could compose separate objects together, but these objects have to be trained using a shared MLP, which limits its flexibility and potential usage.
Yang \textit{et~al.}~\cite{yang2021learning} propose Object-NeRF for editable rendering, while it requires a 2D segmentation model to extract the target object and renders at a slow speed.
% Lazova \textit{et~al.}\cite{lazova2022control} proposed Control-NeRF,  allowing intuitive control and editing, but it could not realize cross-scene composition without retraining.
% Plenoxels\cite{fridovich2022plenoxels} supports composition of different scenes, but suffers from the large storage on the dense index matrix.
Yuan \textit{et~al.} propose NeRF-Editing~\cite{yuan2022nerf}, which allows users to perform controllable shape deformation on the implicit representation.
Yang \textit{et~al.}~\cite{yang2022neumesh} present a mesh-based representation by encoding the implicit field with disentangled geometry and texture codes on mesh vertices, which facilitates a set of editing functionalities.
Tang \textit{et~al.} propose CCNeRF~\cite{tang2022compressible}, which could transform and compose several scenes into one scene with a small model.

\section{Method}
In this section, we first revisit the state-of-the-art point cloud rendering technique and discuss its dilemma of insufficient frequency representation, in Sec.~\ref{subsec:preliminaries}.
In Sec.~\ref{subsec:main_model}, we present our proposed frequency-modulated radiance mapping technique to address above limitation, followed by the introduction to our end-to-end point cloud rendering pipeline in Sec.~\ref{subsec:pipline}.
We also introduce a novel point cloud geometry optimization pipeline as an optional preprocessing module in Sec.~\ref{subsec:pc_opt}.
Applications of the above pipeline on interactive object level editing and multi-scene composition are presented in Sec.~\ref{subsec:edit_model}.

\subsection{Revisiting Point Cloud Rendering}\label{subsec:preliminaries}
 % CR2
Some traditional point cloud rendering methods focus on fast rendering of large-scale point clouds~\cite{schutz2019real, schutz2021rendering}, while
they are not as photorealistic as neural rendering (\emph{e.g.}, not smooth and lack of view-dependent appearance).
Different from traditional works,
our approach builds on a state-of-the-art point-based neural rendering method~\cite{oursaaai}, which utilizes point cloud to determine the position of the surface by the following coordinate rectification
\begin{equation}\label{equ:coor_rect}
\mathbf{x}=\mathbf{o}+\frac{z}{\cos \theta} \times \mathbf{d},
\end{equation}
where $\mathbf{o}$, $\mathbf{d}$ and $\theta$ denote camera position, normalized ray direction and angle between the ray and the $z$-axis pointing to the center of the scene, respectively.
% The depth buffer $z$ is obtained by rasterization.
In practice, $z$ value is obtained by querying the depth buffer during rasterization.
As point clouds are discrete, each point is expanded to a disk of radius $r$ during rasterization~\cite{zwicker2001surface}.
The estimated surface coordinates $\mathbf{x}$ together with ray direction $\mathbf{d}$ are fed to an MLP $\mathbf{F}_{\Theta}$ (${\Theta}$ is the network parameter set) to obtain the features $\mathbf{L}$ which encode the color of the surface point
\begin{equation}
\mathbf{L}=\mathbf{F}_{\Theta}(\mathbf{x}, \mathbf{d}).
\end{equation}
Compared to volume rendering-based models~\cite{mildenhall2021nerf, xu2022point}, 
this method avoids heavy sampling by estimating surface coordinates from point clouds when calculating ray color, and thus achieving real-time rendering. 
As point clouds are usually noisy and the initial depth obtained by rasterization might also be inaccurate, previous methods~\cite{aliev2020neural, rakhimov2022npbg++, oursaaai} often use U-Net structure~\cite{ronneberger2015u} for further refinement.
\begin{figure}[!t]
    \includegraphics[width=\linewidth]{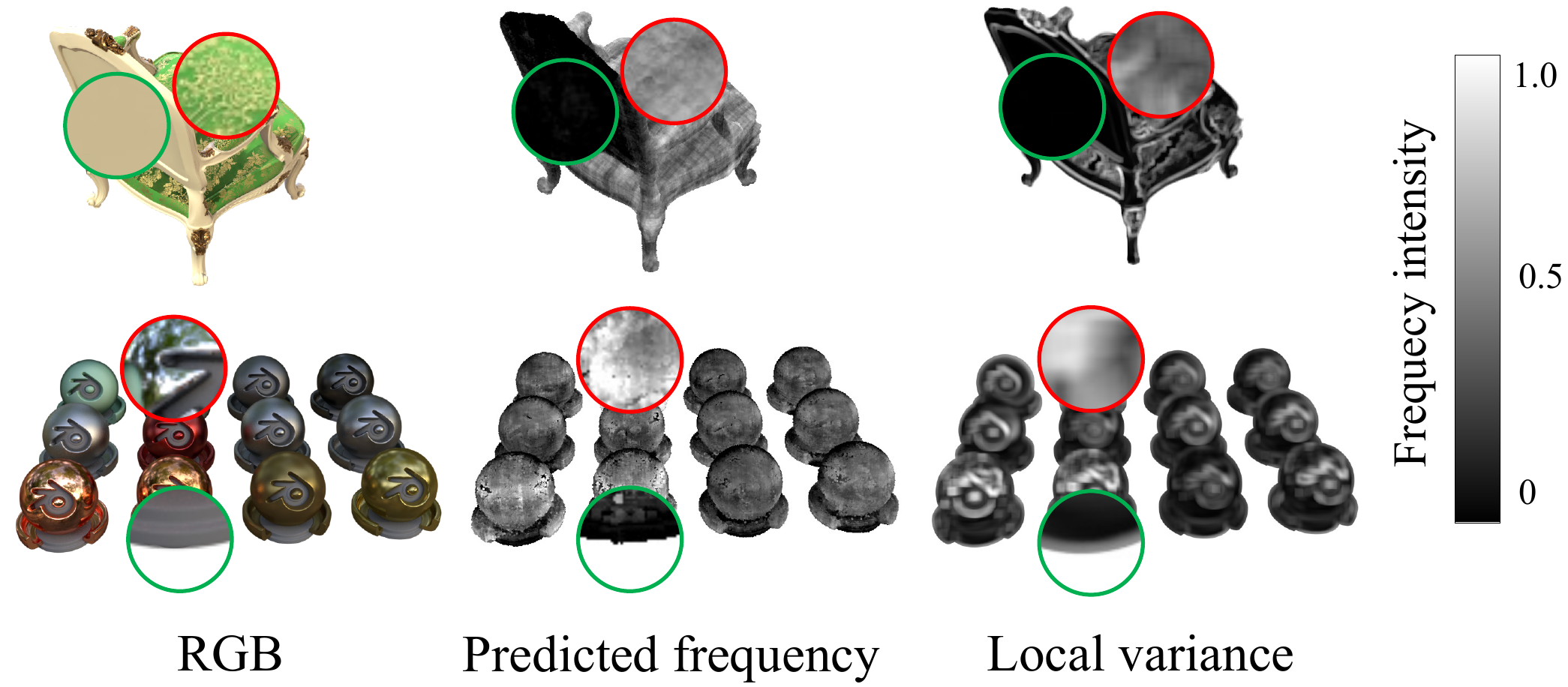}
    \caption{Original RGB images, frequency maps predicted by our hypernetwork, and local variance of original images for reference.
    % The frequency map shows the normalized frequencies predicted by hypernetwork.
    We find that the frequencies are large for strongly textured regions (shown as gray and white color), and the frequencies are close to zeros for weakly textured regions (shown as black color).
    %The local variance map of the RGB image is also given for reference.
    }
    \label{fig:freq_view}
\vspace{-0.15in}
\end{figure}
\vspace{-0.15in}
\paragraph{Limitations.} 
The limitations of the above surface point-based rendering scheme are obvious.
Namely, the point cloud quality is usually poor due to sparsity and large number of outliers, which leads to lower-fidelity rendering especially in high frequency shapes, \emph{i.e.},  highly-3D-textured regions.
Simply increasing the network can not solve the problem.
On the contrary, prevailing volumetric rendering techniques such as NeRF~\cite{mildenhall2021nerf} do not suffer from this issue (\emph{i.e.}, usually with higher rendering quality), since through each sampled ray there exist multiple sampling points near the surface fused together to contribute the target color, which is more stable and sufficient for dealing with high frequency shapes.
Therefore, it is highly demanded to enhance the frequency expression ability of point cloud based implicit surface radiance mapping method at a relatively low computational cost overhead.

\subsection{Frequency-Modulated Radiance Mapping} \label{subsec:main_model}
\begin{figure}[!t]
    \centering
    \includegraphics[width=\linewidth]{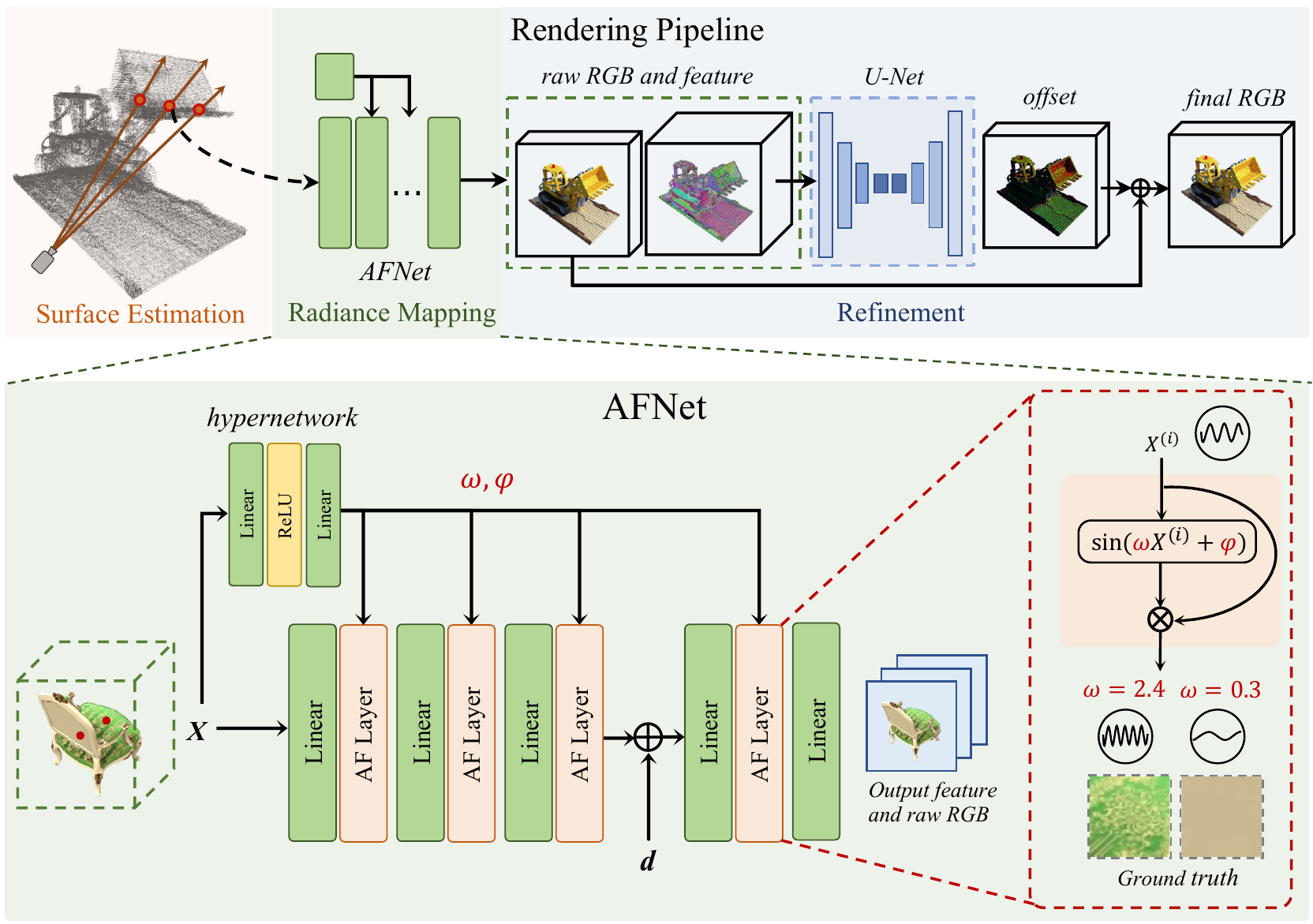}
    \caption{Overview of our rendering pipeline. In the heart of the pipeline is an adaptive frequency modulation module called AFNet, which utilizes a hypernetwork to learn local texture frequency encoding that is consecutively injected to adaptive frequency activation layers to modulate the implicit radiance signal.
}
  \label{fig:fig_model}
  \vspace{-0.2in}
\end{figure}

\paragraph{Motivation.} 
To achieve radiance mapping, also known as radiance field, previous works input a spatial coordinate and infer the color or other features associated with this spatial position via standard MLP with ReLU activation layer.
However, this classical neural network is well-known for its strong spectral bias towards generating lower frequency patterns~\cite{rahaman2019spectral}, which has obvious difficulty in handling high frequency shape details, \emph{e.g.}, 3D textures of objects and occupancy of edges.
Recent works~\cite{tancik2020fourier, sitzmann2020implicit, fathony2020multiplicative} analyze the expressive power and inductive bias of the network and enhance the fitting ability of the network from a global perspective, while we emphasize the local perspective of the problem, \emph{i.e.}, the frequencies of a 3D scene are region-dependent.
Namely, most real objects are composed by both weak textured regions and strong textured ones.
For example, in the Chair scene of NeRF-Synthetic dataset, the back of the chair belongs to  weak texture areas, and the cushion with patterns belongs to strong texture areas, as shown in Fig.~\ref{fig:freq_view}.
If the signal frequency does not match the target area, the implicit radiance mapping would be poor due to misaligned frequency functional support.
Thus, the above observations motivate us to propose an Adaptive Frequency Net (AFNet) for frequency-modulated radiance mapping, as shown in Fig.~\ref{fig:fig_model}, which is capable of selecting high frequency primitives for modeling rich-3D-textured regions while uses low frequency basis for presenting non-textured regions, in order to yield overall high-fidelity rendering quality.

\vspace{-0.15in}

\paragraph{Adaptive Frequency Estimation and Modulation.}
The major working mechanism of the proposed AFNet is to first estimate the local frequency information for each point, and then this frequency encoding is injected to the proposed Adaptive Frequency Activation Layers (AF Layers) which adaptively select a proper set of frequency component, providing a more precise support for generating high-fidelity reconstruction, as shown in Fig.~\ref{fig:fig_model}.
Therefore, the core building block of the proposed AFNet includes 1) AF Layers and 2) a hypernetwork for local frequency estimation, which are introduced in detail as follows.

\vspace{-0.15in}

\paragraph{Adaptive Frequency Activation Layer.}
% a feature generation module
Assuming local frequency information is successfully estimated and encoded as scalar frequencies and phases, our AF layers first encode the input radiance signal with a nonlinear filter (we use a sinusoidal filter) parameterized by the obtained point-wise adaptive frequency and phase. 
Then, a Hadamard product between the filtered feature and the input feature is taken as the activated output, which could be regarded as a frequency modulation on the input signal. 
Mathematically, suppose that the input feature of $i$-th layer is $\mathbf{X}^{(i)} \in \mathbb{R}^{M\times C}$, in which $M$ and $C$ are point number and feature dimension respectively, and the adaptive frequencies and phases are $\mathbf{\omega}^{(i)}, \mathbf{\phi}^{(i)} \in \mathbb{R}^{M\times 1}$, whose dimensions will be broadcast to $\mathbb{R}^{M\times C}$ during computation. Then the adaptive frequency activation layers can be expressed as:
\begin{equation}\label{equ:3}
    \mathbf{Y}^{(i)} = \mathbf{X}^{(i)} \circ \sin(\omega^{(i)}\circ\mathbf{X}^{(i)} + \phi^{(i)}).
\end{equation}
% CR1
The design of Eqn.~\ref{equ:3} has a theoretical justification.
We measure the local frequency by the derivative of the color with respect to the spatial coordinate, and adaptively adjust the derivative by multiplying the feature $\mathbf{X}^{(i)}$ by the predicted frequency $\omega^{(i)}$. 
We then use the sine function to constrain the numerical range. The final form, $\mathbf{X}^{(i)}\circ\sin(\cdot)$, is borrowed from MFN~\cite{fathony2020multiplicative}.

\vspace{-0.15in}
\paragraph{Hypernetwork for Frequency Estimation.}
The adaptive frequencies and phases are expected to contain important information of the 3D texture variation on the surface position; however, to obtain the accurate frequency representation of each point is non-trivial.
Inspired by HyperNetwork~\cite{ha2016hypernetworks}, an approach of using a small network to generate the weights for a larger network, we use a two-layer MLP as hypernetwork to infer the frequency and phase parameters from each spatial point. 
In Fig.~\ref{fig:freq_view}, we show the predicted frequency of the 4-th adaptive frequency activation layer.
As seen in the RGB images, the back of the chair and the base of the ball have weak texture, while the frequency predicted by hypernetwork is closed to 0 (shown as black color).
For strongly textured regions, such as patterned cushions in chair and the metal material, the hypernetwork predicts large frequency (shown as grey and white).
These observations empirically indicates that the proposed hypernetwork is capable of learning correct frequency information even there is \textbf{NO} supervision on frequency label.

\subsection{End-to-end Rendering Pipeline} \label{subsec:pipline}
% \paragraph{Overview.} 
Our pipeline consists of three stages: surface estimation, radiance mapping and refinement, as shown in Fig.~\ref{fig:fig_model}.
% In surface estimation stage, we first rasterize the point cloud to obtain the depth map and use equation \ref{equ:coor_rect} to estimate the coordinate of the surface where each ray intersects. 
% \paragraph{Surface estimation.} 
In the surface estimation stage, we first rasterize the point cloud $\mathbf{P} = \{\mathbf{p}_1, \mathbf{p}_2, \cdots, \mathbf{p}_N\} \in \mathbb{R}^{N\times3}$ to obtain the depth buffer $\mathbf{z}\in\mathbb{R}^{H\times W}$ at a given view.
Specifically, we transform the point cloud from the world coordinate space to the normalized device coordinate sapce, in which $\mathbf{p}_k  \in [-1,1] \times [-1,1]\times[z_{near}, z_{far}], k=1,2,...,N$.
Similarly, we transform the pixel coordinates $\mathbf{I} \in \mathbb{R}^{H\times W \times 2}$ to normalized space such that $\mathbf{I}_{i,j} \in [-1,1]^2$.
Then, each point $\mathbf{p}_k$ is expanded to a disk of radius $r_k$ in $xOy$ plane. The depth buffer $\mathbf{z}_{i,j}$ is the depth of the nearest disk covering pixel $\mathbf{I}_{i,j}$, that is
\begin{equation}
    \mathbf{z}_{i,j} = \min_k{\mathbf{p}^z_k}, ~~~~ s.t.~ (\mathbf{p}_k^x - \mathbf{I}_{i,j}^x)^2 + (\mathbf{p}_k^y - \mathbf{I}_{i,j}^y)^2 < r_k^2,
    \label{equ:surf_est}
\end{equation}
which can be seen as a special case of surface splatting~\cite{zwicker2001surface}.
Then, we use Eqn.~\ref{equ:coor_rect} to estimate the coordinate of the surface where each ray intersects. 
\begin{figure}[!t]
    \centering
    \includegraphics[width=\linewidth]{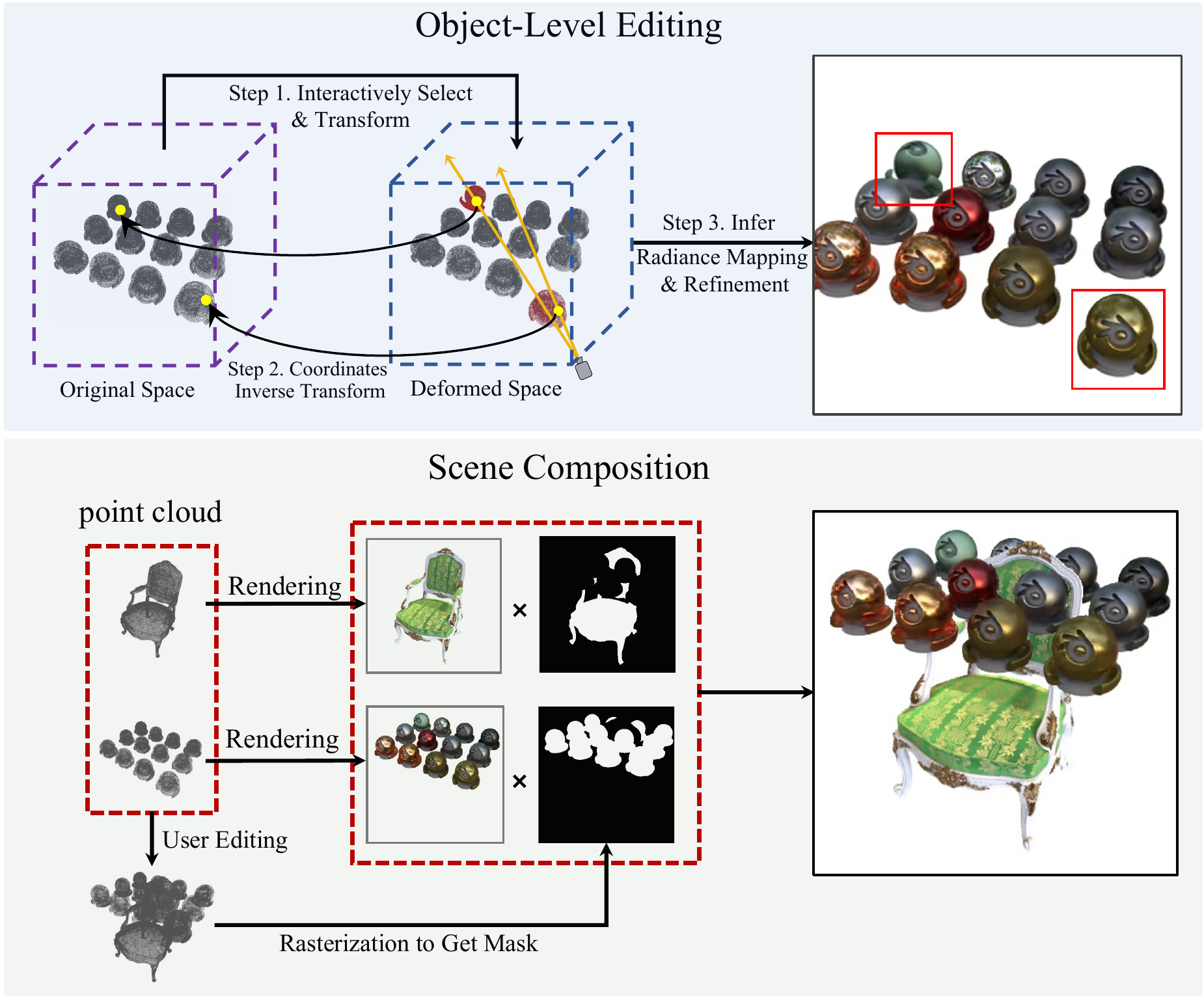}
    \vspace{-0.25in}
    \caption{Object-level editing and scene composition pipelines.}
  \label{fig:fig_edit}
  \vspace{-0.2in}
\end{figure}

In radiance mapping stage, the estimated surface coordinates are fed into the proposed AFNet to predict the raw RGB values and feature vectors. In the refinement stage, the feature vectors will be fed into an U-Net to predict the offsets. The final RGBs are obtained by adding the raw RGB values and offsets.
AFNet and U-Net are trained with ground truth images as supervision by using L2 loss and perceptual loss~\cite{zhang2018perceptual}.

\subsection{Preprocessing: Geometry Optimization}\label{subsec:pc_opt} 
% \vspace{-0.05in}
In the meantime, point clouds obtained by MVS method~\cite{yao2018mvsnet} or 3D scans inevitably contain noise or missing parts of the surface, which harms the rendering quality or even produces artifacts.
In view of this, we design a point cloud geometry optimization strategy consisting of a point cloud denoising step and a completion step, considered as an optional preprocessing module in our framework, to further enhance rendering quality.

\vspace{-0.15in}
\paragraph{Denoising.}
We design a volume rendering-based pipeline specifically to optimize the point cloud geometry.
In rasterization, we maintain the depth buffer of $k$ points instead of the nearest point. The depth values are used to compute the rectified coordinates of the points using Eqn.~\ref{equ:coor_rect}. The rectifed coordinates, combined with ray directions, are fed into a MLP to estimate colors. 
Besides, an optimizable opacity parameter $\alpha_i$ is assigned to each point $\mathbf{p}_i$. 
Pixel RGB values are obtained using the volume rendering equation. Inspired by Point-NeRF~\cite{xu2022point}, we train the MLP and the opacity parameters to fit the ground-truth images with the following sparse regularization:
\vspace{-0.1in}
\begin{equation}
\mathcal{L}_{\text{sparse }}=\frac{1}{N} \sum_{i=1}^N\left[\log \left(\alpha_i\right)+\log \left(1-\alpha_i\right)\right].
\end{equation}
% \vspace{-0.1in}
When the training converges, we remove those low-opacity points with a threshold $\mu$.

\vspace{-0.15in}

\paragraph{Completion.}
For each view in training set, if a pixel in the ground truth image is occupied, but the buffer obtained by rasterizaion is empty, we add a set of points with opacity parameters along the pixel ray in the point clouds.
% For those pixels which have empty buffer in rasterization, we add a set of points with density parameters along the projected ray in the point clouds.
To determine the depth range of the added points, we refers to the depth range of adjacent pixel buffer.
Although the completion step introduces additional noise, it would be removed in the denoising stage of the next iteration.

Above denoising and completion steps run alternatively until convergence.
Please see supplementary material for more details.

% \vspace{-0.15in}
\subsection{Scene Editing}\label{subsec:edit_model}
In this section, we introduce an editing method based on the proposed rendering pipeline. 
At object level, we leverage point clouds to bridge the gap between user editing and deformation field.
At scene level, the proposed method can combine different scenes without cross-scene training.

% \vspace{-0.15in}

\begin{table*}[!t]
    \small
    \centering
    \setlength\tabcolsep{1pt}
    % \footnotesize
    \vspace{-8pt}
    \begin{tabular*}{\hsize}{@{}@{\extracolsep{\fill}}lcccccccccccc@{}}
    \toprule
     & \multicolumn{3}{c}{NeRF-Synthetic} & \multicolumn{3}{c}{Tanks\&Temples }  & \multicolumn{3}{c}{ScanNet} & \multicolumn{3}{c}{DTU} \\
     Method & 
     \small{PSNR$\uparrow$}&  \small{SSIM$\uparrow$} &  \small{LPIPS$\downarrow$} & \small{PSNR$\uparrow$}&  \small{SSIM$\uparrow$} &  \small{LPIPS$\downarrow$} & \small{PSNR$\uparrow$}&  \small{SSIM$\uparrow$} &  \small{LPIPS$\downarrow$} & \small{PSNR$\uparrow$}&  \small{SSIM$\uparrow$} &  \small{LPIPS$\downarrow$} \\
    \midrule
    NeRF~\cite{mildenhall2021nerf} & 31.01 & 0.947 & 0.081 & 25.78 & 0.864 & 0.198 & 25.74 & 0.780 & 0.537 & 26.92 & 0.909 & 0.198\\
    CCNeRF-CP~\cite{tang2022compressible} & 30.55 & 0.935 & 0.076 & 27.01 & 0.879 & 0.180 & 24.65 & 0.774& 0.542& 26.79& 0.907& 0.178 \\
    \small{CCNeRF-HY-S}~\cite{tang2022compressible} & 31.22 & 0.947 & 0.074 & 27.53 & 0.901 & 0.177 & 25.17 & 0.781& 0.539& 27.23& 0.910& 0.171\\
    
    \midrule
    NPBG~\cite{aliev2020neural} & 28.10 & 0.923 & 0.077 & 25.97 & 0.889 &0.137 &  25.09 & 0.737 & 0.459 & 26.00 & 0.895 & 0.130 \\
    NPBG++~\cite{rakhimov2022npbg++}& 28.12 & 0.928 & 0.076& 26.04 & 0.892 & 0.130 & 25.27 & 0.772 & 0.448 & 26.08 & 0.895 & 0.131\\
    Huang \textit{et~al.}~\cite{oursaaai} & 28.96 & 0.932 & 0.061 & 26.35 & 0.893 & 0.130 & 25.88 & 0.794 & 0.415 & 26.22 & 0.900 & 0.132\\
    Ours & \textbf{31.24} & \textbf{0.950} & \textbf{0.049} & \textbf{27.79} & \textbf{0.902} & \textbf{0.125} & \textbf{26.66} & \textbf{0.803} & \textbf{0.400} & \textbf{27.27} & \textbf{0.908} & \textbf{0.129}\\
    
    \bottomrule
    \end{tabular*}
    \vspace{-0.1in}
    \caption{Quantitative evaluation on NeRF-Synthetic~\cite{mildenhall2021nerf}, Tanks and Temples~\cite{knapitsch2017tanks}, ScanNet~\cite{dai2017scannet} and DTU~\cite{jensen2014large} datasets. 
    Since our work focuses on surface point-based method, which supports real-time and editable rendering, we mainly compare with NPBG~\cite{aliev2020neural}, NPBG++~\cite{rakhimov2022npbg++} and Huang \textit{et al.}~\cite{oursaaai}. We also 
   show the performence of 
   NeRF~\cite{mildenhall2021nerf} and CCNeRF~\cite{tang2022compressible} for reference.}
    \label{tab:benchmark}
    % \vspace{-10pt}
    \vspace{-0.1in}
\end{table*}

\vspace{-0.15in}
\paragraph{Object-level Editing.}
As an explicit 3D representation, point clouds are ideal for interactive editing.
After editing target points, we can estimate the new surface coordinates of the edited point cloud by rasterization. 
Since radiance mapping encodes appearance of the surface in  original 3D space, 
we need to transform the estimated surface points to original encoding space, which can be regarded as a deformation of the original space, \emph{i.e.}, deformation field~\cite{park2021nerfies,pumarola2021d}.
Specifically, we keep the indices of the points in pixel buffer during rasterization and query the transformations performed on those points. Then we perform the corresponding inverse transformation on the estimated surface coordinates to obtain the locations in original 3D space, which replaces the calculation of deformation field, as shwon in Fig.~\ref{fig:fig_edit}. Finally, we feed the transformed surface points to radiance mapping module followed by refinement module to obtain the rendering result.

\vspace{-0.15in}
\paragraph{Scene Composition.}
A direct way to render a combined point cloud is to use respective AFNets for radiance mapping, and a shared U-Net for refinement.
However, due to the inductive bias of U-Net, the texture of different scenes will affect each other, which would reduce the rendering quality. Moreover, cross-scene training of U-Net impairs the flexibility of editing, \emph{i.e.}, adding a new scene to an existing composition requires retraining together.
To solve the above problems, we compose the scenes in pixel level. As shown in Fig.~\ref{fig:fig_edit}, we first rasterize the combined point clouds to determine which scene each pixel belongs to and generate a mask for each scene. Then we render each scene separately and multiply the masks with the rendered images. We combine the masked images to obtain the final result. 
% Please see supplementary material for more details of scene composition.

\begin{figure}[!t]
\vspace{-8pt}
    \centering
    % \fbox{\rule{0pt}{4in} \rule{.9\linewidth}{0pt}}
    \includegraphics[width=\linewidth]{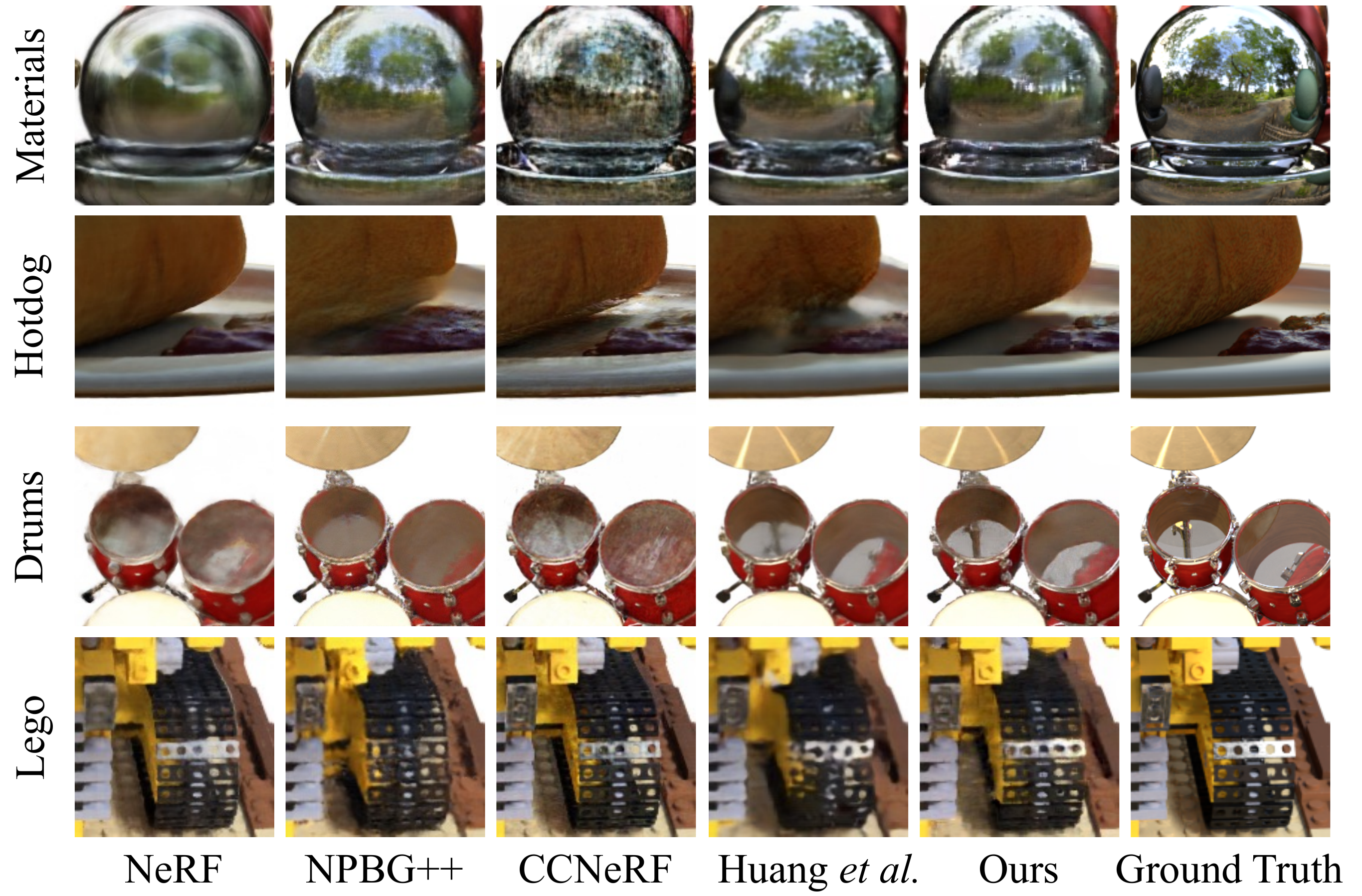}
    \caption{Qualitative evaluations on NeRF-Synthetic~\cite{mildenhall2021nerf} dataset. Our method renders sharper detailed textures without artifacts.}
  \label{fig:benchmark_nerf}
 \vspace{-0.2in}
\end{figure}

\section{Experiments}
\subsection{Experimental Settings}
\paragraph{Datasets and Compared Methods.}
Since our work focuses on surface point-based rendering methods, which enable real-time and editable rendering, we mainly compare with the following methods:
% \begin{itemize}
1) NPBG~\cite{aliev2020neural}: A famous surface point-based rendering method which enables real-time rendering and editing; 2) NPBG++~\cite{rakhimov2022npbg++}: The improved version of NPBG; and 3) Huang \textit{et al}.~\cite{oursaaai}: The state-of-the-art surface point-based rendering pipeline.
% \end{itemize}
We first compare our performance with the above methods on NeRF-Synthetic~\cite{mildenhall2021nerf}, Tanks and Temples~\cite{knapitsch2017tanks}, ScanNet~\cite{dai2017scannet} and DTU~\cite{jensen2014large} datasets.
% We experiment on four datasets: NeRF-Synthetic\cite{mildenhall2021nerf}, Tanks and Temples\cite{knapitsch2017tanks}, ScanNet\cite{dai2017scannet} and DTU\cite{jensen2014large}.
To further demonstrate the superiority of our method, we conduct a comprehensive comparison with volume rendering-based methods, including {NeRF}~\cite{mildenhall2021nerf} and its variants, on NeRF-Synthetic~\cite{mildenhall2021nerf} and Tanks and Temples~\cite{knapitsch2017tanks} datasets.
% We give a detailed description on those datasets and methods in supplementary material. 
Please see supplementary material for more details.

\begin{table*}[!ht]
\small
\centering
    \begin{tabular}{l|c|c|cc|ccc|ccc}
        \hline
        \multicolumn{3}{c}{~} &\multicolumn{2}{|c|}{\small{Editing Ability}} & \multicolumn{3}{|c|}{NeRF-Synthetic}& \multicolumn{3}{|c}{Tanks\&Temples}\\ 
        \hline
        Method & \small{Size(MB)}&  \small{FPS}& \small{Object} & \small{Scene}&  \small{PSNR$\uparrow$}&  \small{SSIM$\uparrow$} &  \small{LPIPS$\downarrow$} &\small{PSNR$\uparrow$}&  \small{SSIM$\uparrow$} &  \small{LPIPS$\downarrow$}  \\
         
        \hline
        % SRN & - & 0.909 & \ding{55}& \ding{55}& 22.26 & 0.846 & 0.170 & 24.10 & 0.847 & 0.251  \\ % from plenoctrees
        NeRF~\cite{mildenhall2021nerf} & 5.0 & 0.023 & \color{BrickRed}{\ding{55}}& \color{BrickRed}{\ding{55}}& 31.01 & 0.947 & 0.081 & 25.78 & 0.864 & 0.198 \\
        NSVF~\cite{liu2020neural} & 16.0 & 0.815 & \color{BrickRed}{\ding{55}}&  \color{ForestGreen}{\ding{51}}& 31.75 & 0.953 & 0.047 & 28.40 & 0.900 & 0.153  \\
        Object-NeRF~\cite{yang2021learning} &121.2 & 0.1 &  \color{ForestGreen}{\ding{51}}& \color{BrickRed}{\ding{55}}& 31.19 & 0.949 & 0.079 & 25.96 & 0.866 & 0.194   \\
        PlenOctrees~\cite{yu2021plenoctrees} & 1976.3 & 168 & \color{BrickRed}{\ding{55}}& \color{ForestGreen}{\ding{51}}& 31.71 & 0.958 & 0.053 & 27.99 &0.917 & 0.131  \\
        Point-NeRF~\cite{xu2022point} & 20.0 & 0.125 & \color{BrickRed}{\ding{55}}& \color{BrickRed}{\ding{55}} & 33.00 & 0.978 & 0.055 & 29.61 &0.954 & 0.115  \\
        Plenoxels~\cite{fridovich2022plenoxels}& 778.1 & 15 & \color{BrickRed}{\ding{55}}&  \color{ForestGreen}{\ding{51}}& 31.71  & 0.958 & 0.049 & 27.43 & 0.906 & 0.142  \\
        TensoRF~\cite{chen2022tensorf} & 71.8 & 1.15 & \color{BrickRed}{\ding{55}}& \color{ForestGreen}{\ding{51}}& 33.14 & 0.963 & 0.056&28.56 &0.920&  0.118 \\
        Instant-NGP~\cite{muller2022instant}  & 63.3 & 60 & \color{BrickRed}{\ding{55}}& \color{BrickRed}{\ding{55}}& 33.18 & 0.963 & 0.050&28.78&0.925&0.113 \\
        % DFF & & & \ding{55}& \ding{55}& & &   \\
        \small{CCNeRF-CP}~\cite{tang2022compressible} & 4.4 & 1.05 & \color{BrickRed}{\ding{55}}&  \color{ForestGreen}{\ding{51}}
        & 30.55 & 0.935 & 0.076 & 27.01 & 0.879 & 0.180 \\
        \small{CCNeRF-HY-S}~\cite{tang2022compressible}& 68.9 &1.05 & \color{BrickRed}{\ding{55}}&  \color{ForestGreen}{\ding{51}}& 31.22 & 0.947 & 0.074 & 27.53 & 0.901 & 0.177  \\
        \hline
        NPBG~\cite{aliev2020neural} & 44.2 & 33.4 &  \color{ForestGreen}{\ding{51}}&  \color{ForestGreen}{\ding{51}} & 28.10 & 0.923 & 0.077 & 25.97 & 0.889 &0.137  \\
        NPBG++~\cite{rakhimov2022npbg++}& 28.6 & 35.4 &  \color{ForestGreen}{\ding{51}}&  \color{ForestGreen}{\ding{51}}& 28.12 & 0.928 & 0.076& 26.04 & 0.892 & 0.130   \\
        % DPRF  & & & \ding{55}& \ding{55}& & &   \\
        Huang~\textit{et~al.}\cite{oursaaai}  & 18.5 & 39.1 &  \color{ForestGreen}{\ding{51}}&  \color{ForestGreen}{\ding{51}}& 28.96 & 0.932 & 0.061 & 26.35 & 0.893 & 0.130   \\
        Ours  & 11.8 & 39.3  & \color{ForestGreen}{\ding{51}}&  \color{ForestGreen}{\ding{51}}& {31.24} & {0.950} & {0.049} & {27.79} & {0.902} & {0.125}   \\
        \hline
    \end{tabular}
    \vspace{-0.1in}
    \caption{\label{tab:all}Comprehensive evaluation. We report the model size (including point cloud size for point-based methods), FPS on NeRF-Synthetic\cite{mildenhall2021nerf} dataset with 800$\times$800 resolution, editing ablity of object-level transform and scene composition, and performance on NeRF-Synthetic\cite{mildenhall2021nerf} and Tanks and temples\cite{knapitsch2017tanks} datasets. The results demonstrate the excellent comprehensiveness of our method.}
    \vspace{-0.1in}
    
\end{table*}

\begin{figure}[!t]
\vspace{-0.1in}
    \centering
    % \fbox{\rule{0pt}{4in} \rule{.9\linewidth}{0pt}}
    \includegraphics[width=\linewidth]{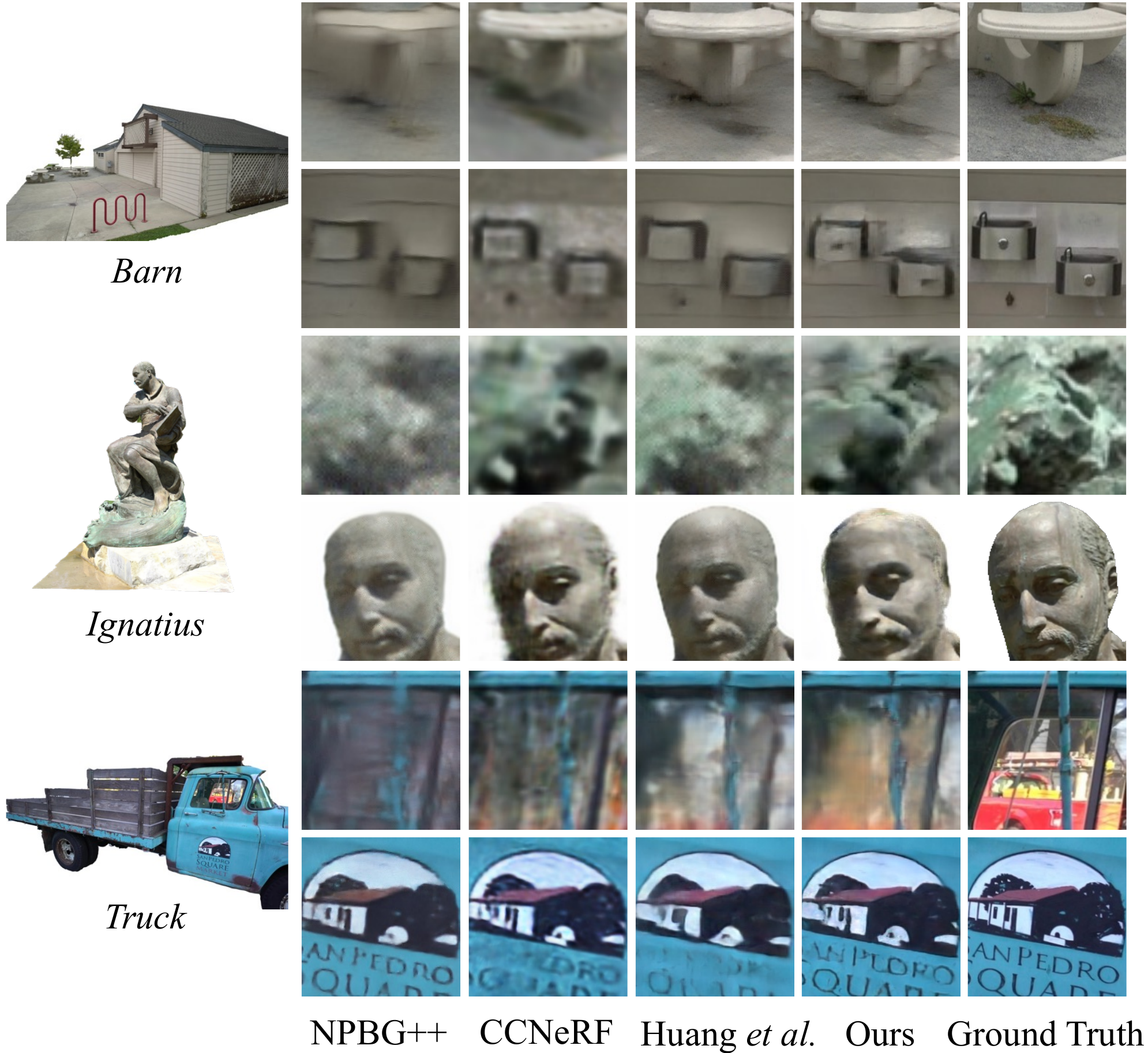}
    \vspace{-0.2in}
    \caption{Qualitative evaluations on Tanks and Temples~\cite{knapitsch2017tanks} dataset. Our method renders sharper detailed textures.}
  \label{fig:benchmark_tt_dtu_scan}
 \vspace{-0.25in}
\end{figure}

\vspace{-0.15in}
\paragraph{Implementation Details.}
There are two ways of rasterization: PyTorch3D\cite{ravi2020pytorch3d} for headless server and OpenGL\cite{shreiner2009opengl} for real-time rendering.
Since our training is performed on a headless server, our main results are based on PyTorch3D rasterization.
The widths of AFNet and hypernetwork are both 256, and camera ray directions are concated to the   third layer output feature of AFNet. 
Moreover, we modify the positional encoding of NeRF\cite{mildenhall2021nerf} to make it more suitable for modulation, which is described in detail in supplementary material.
The architecture of our U-Net is consistent with NPBG~\cite{aliev2020neural}, \emph{i.e.}, gated blocks are used, consisting of gated convolution and instance normalization.
The U-Net down-samples the feature map for four times and the dimensions of each layer are 16, 32, 64, 128, 256. 
We train the AFNet and U-Net with L2 loss and perceptual loss\cite{zhang2018perceptual} on a single NVIDIA GeForce RTX 3090 GPU.
For preprocessing of point cloud optimization, we use only L2 loss and a 5-layers MLP with 256 width, where ray directions are concated to the third layer output feature. 
We use Adam optimizer for training, with a batch size of 2. The initial learning rates of AFNet and U-Net are 5e-4 and 1.5e-4 respectively. 

\subsection{Results of View Synthesis}
We present quantitative comparison results on three performance metrics (PSNR, SSIM and LPIPS) in Tab.~\ref{tab:benchmark}. As can be seen, we outperform the previous point-based methods with a significant margin on all datasets. 
Our method outperforms NeRF~\cite{mildenhall2021nerf} and is comparable to the high-profile version of CCNeRF~\cite{tang2022compressible}.
Figs.~\ref{fig:benchmark_nerf} and~\ref{fig:benchmark_tt_dtu_scan} present a qualitative comparison on NeRF-Synthetic~\cite{mildenhall2021nerf} and Tanks and Temples~\cite{knapitsch2017tanks} dataset. As can be seen, our method produces sharper details without introducing artifacts, especially for highly textured regions.

\vspace{-0.15in}
\paragraph{Comprehensive Analysis.}
To further illustrate the superiority of our method, we expand the quantitative experiments with more comparison works in Tab.~\ref{tab:all}, and comprehensively compare them from four aspects: model size, rendering speed, editing ability, and performance.
It can be seen that almost all the follow-up works of NeRF are difficult to take into account the four aspects at the same time. 
PlenOctrees~\cite{yu2021plenoctrees} achieves the state-of-the-art FPS, but its application scenarios are limited due to its huge model size. 
Instant-NGP~\cite{muller2022instant} has high performance and real-time rendering speed, but lacks editing abilities. Object-NeRF~\cite{yang2021learning} and CCNeRF~\cite{tang2022compressible} can achieve object-level editing and scene composition respectively, but cannot achieve real-time rendering. 
Our method achieves high performance and flexible editing with a small model size while rendering in real-time.

\begin{figure}[!htbp]
% \vspace{-0.15in}
    \centering
    \includegraphics[width=\linewidth]{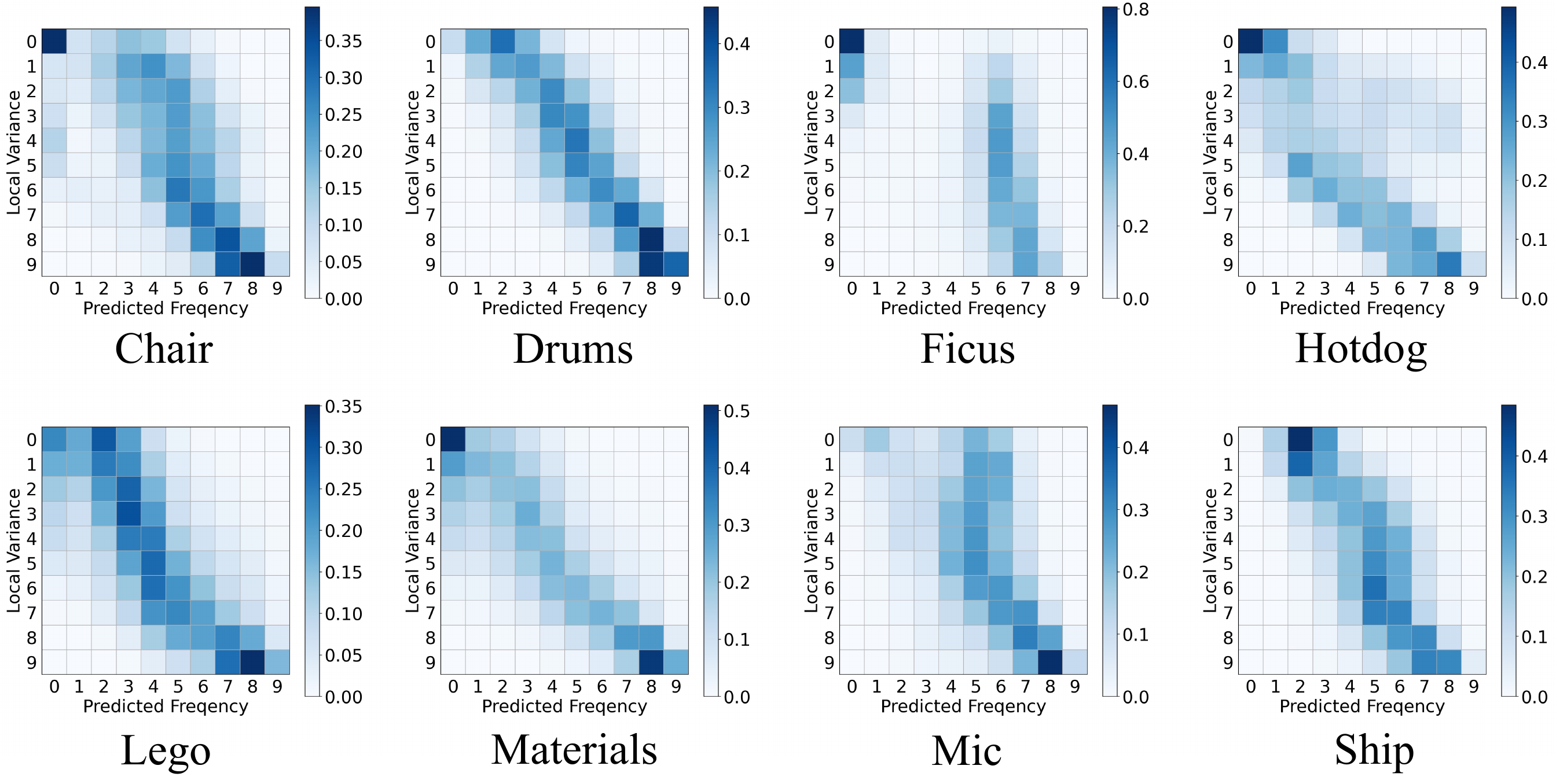}
    \vspace{-0.2in}
    \caption{Confusion matrix between predicted frequency map and actual local variance on 200 test views of NeRF-Synthetic dataset. The frequency interval is divided into 10 segments, and from 0 to 9 are low frequency to high frequency. 
    The frequencies predicted by our hypernetwork are highly correlated with the true frequencies reflected by local variance.}
  \label{fig:ex_freq}
  \vspace{-0.25in}
\end{figure}
\vspace{-0.15in}

\begin{figure}[!t]
    \centering
    \includegraphics[width=\linewidth]{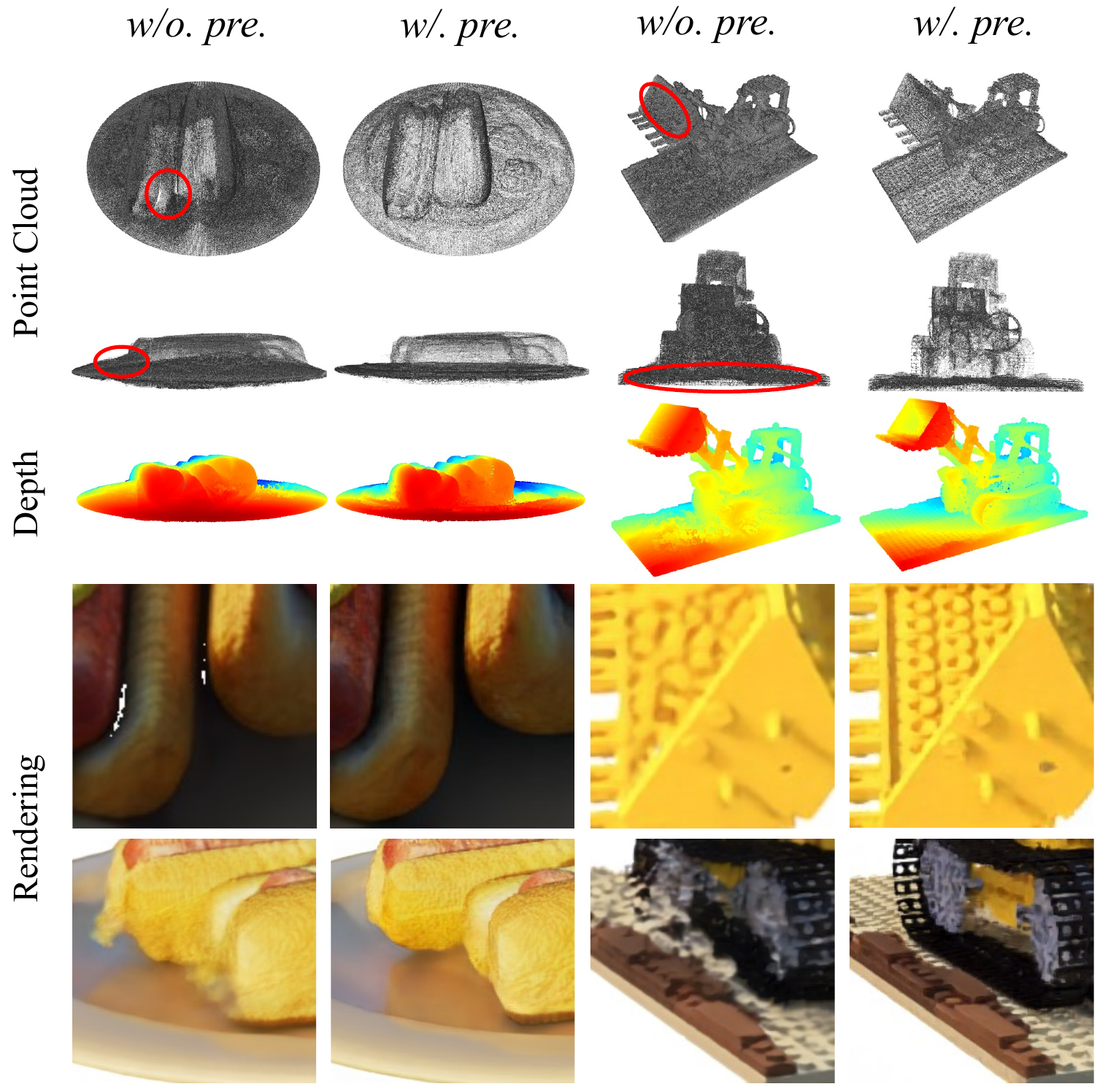}
    \caption{We show the point clouds, depth maps and rendered images with and without preprocessing of point cloud geometry optimization (\textit{w/. pre.} and \textit{w/o. pre.}). The noise near the crawler and the shovel in Lego scene causes artifacts and the missing part of Hotdog causes some blanks in the rendering result, while our preprocessing module optimizes the point cloud geometry and remove those artifacts.}
  \label{fig:ablation_pc_opt}
  \vspace{-0.2in}
\end{figure}

\paragraph{Ablation Study.}
The frequencies predicted by our hypernetwork are highly correlated with the true frequencies reflected by local variance, as shown in Fig.~\ref{fig:ex_freq}.
Note that there is a reasonable gap between our predicted 3D texture frequencies and 2D image frequencies.
For example, in Ficus scene, the occlusion between leaves and background produces a high frequency region in 2D image, but the 3D texture of leaves is weak (just green).
Moreover, we claim that the frequencies predicted by hypernetwork are adaptive, which can modulate the radiance signal better than actual local variance.
To illustrate this, we inject the local variance of 2D images into AF layers without hypernetwork learning. The evaluation results in Tab.~\ref{tab:abla} show that the actual local variance is difficult to modulate the radiance signal.
Also, point cloud geometry optimization is critical for some poorly reconstructed scenes. 
In Tab.~\ref{tab:abla}, we show the quantitative results on PSNR and point scale. 
The preprocessing module improves performance while significantly reducing the size of the point cloud.
In Fig.~\ref{fig:ablation_pc_opt}, we show the point clouds, depth maps and rendered images of Lego and Hotdog scenes. 
As can be seen, the noises near the crawler and the shovel in Lego scene cause artifacts and the missing part of Hotdog causes some blanks in the rendering result, while our preprocessing module optimizes point cloud geometry and removes those artifacts.

\begin{table}[!t]
    \centering
    \small
    \setlength\tabcolsep{1pt}
    % \footnotesize
    \vspace{-8pt}
    \begin{tabular*}{\hsize}{@{}@{\extracolsep{\fill}}lcccccccc@{}}
    \toprule
     & \small{Chair} & \small{Drums} & \small{Ficus} & \small{Hotdog} 
     & \small{Lego} & \small{Materi} & \small{Mic} & \small{Ship} \\
     \midrule
     \multicolumn{9}{c}{{PSNR}}\\
    \midrule
     \textit{loc. var}.& 30.25 &	23.48 &	29.90 &	30.33 &	29.49 &	26.55 &	32.22 &	24.47	 \\
    \midrule 
    \textit{w/o. pre.}  & \small{32.46} & \small{25.15}& \small{30.95}& \small{33.10}& \small{27.51}& \small{28.96}& \small{33.35}& \small{26.78} \\
    {ours} & 33.06 & 25.95 & 32.19 & 35.82 & 31.56 & 29.69 & 33.64 & 27.97  \\
    \midrule
    \multicolumn{9}{c}{Number of points (Millions)}\\
    \midrule
    \textit{w/o. pre.}& 0.69 &	0.54 &	0.40 &	1.02 &	1.07 &	1.19 &	0.50 &	1.52 \\
    {ours} &0.33 &	0.17 &	0.10 &	0.28 &	0.22 &	0.28 &	0.36 &	0.33 \\
    \bottomrule
    \end{tabular*}
    \vspace{-0.1in}
    \caption{Ablation analysis of adaptive frequency and point cloud geometry optimization. We inject the local variance extracted from the ground truth images as frequency information without using hypernetwork to predict frequency, denoted as \textit{loc. var.} We also show the PSNR of training without preprocessing of point cloud geometry optimization (\textit{w/o. pre.}) and the number of points before and after preprocessing.}
    \label{tab:abla}
    \vspace{-0.1in}
\end{table}

\begin{figure}[!htbp]
\vspace{-8pt}
    \centering
    % \fbox{\rule{0pt}{5in} \rule{.9\linewidth}{0pt}}
    \includegraphics[width=\linewidth]{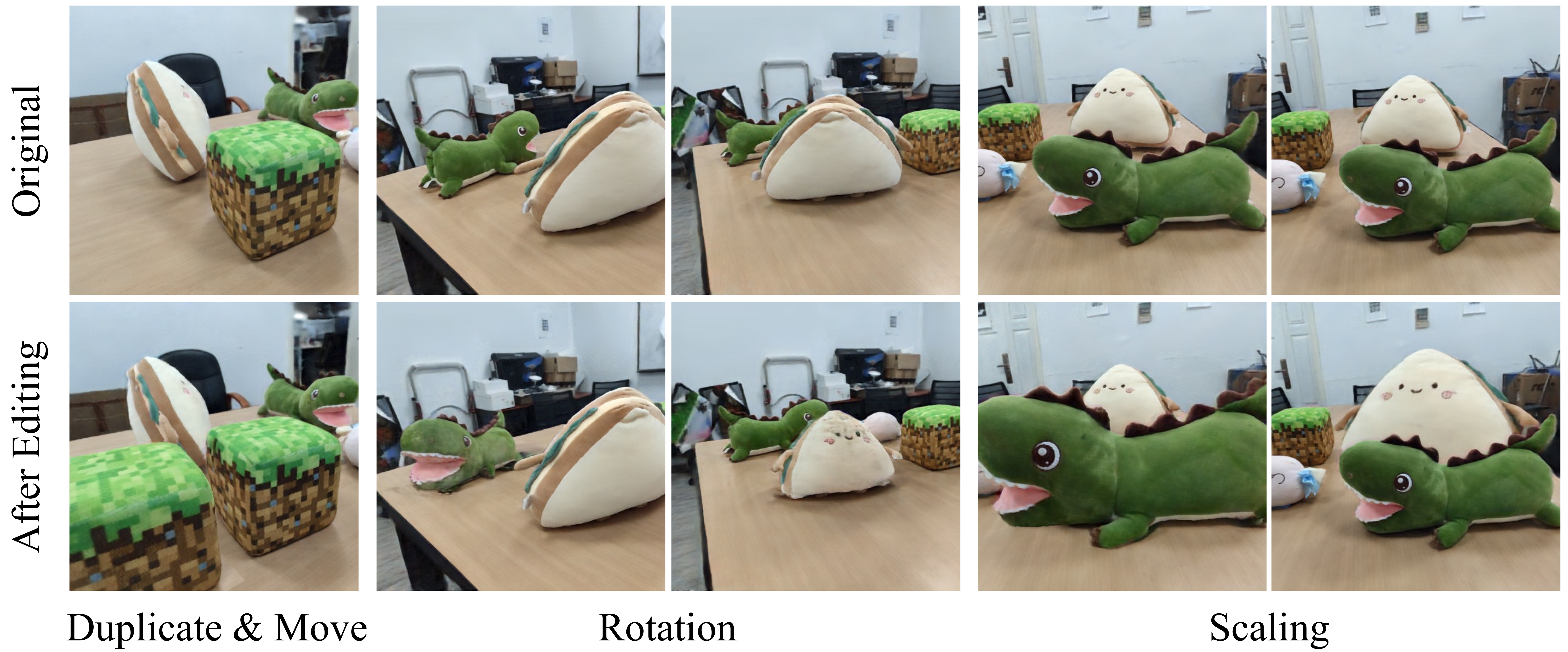}
    \vspace{-0.2in}
    \caption{Object-level editing on ToyDesk~\cite{yang2021learning} dataset. We can scale, duplicate, translate and rotate the objects selected by users.}
  \label{fig:edit_toy}
%   \vspace{-0.1in}
\end{figure}

\subsection{Results of Editing}
\begin{figure}[!t]
% \vspace{-0.15in}
    \centering
    % \fbox{\rule{0pt}{5in} \rule{.9\linewidth}{0pt}}
    \includegraphics[width=3in]{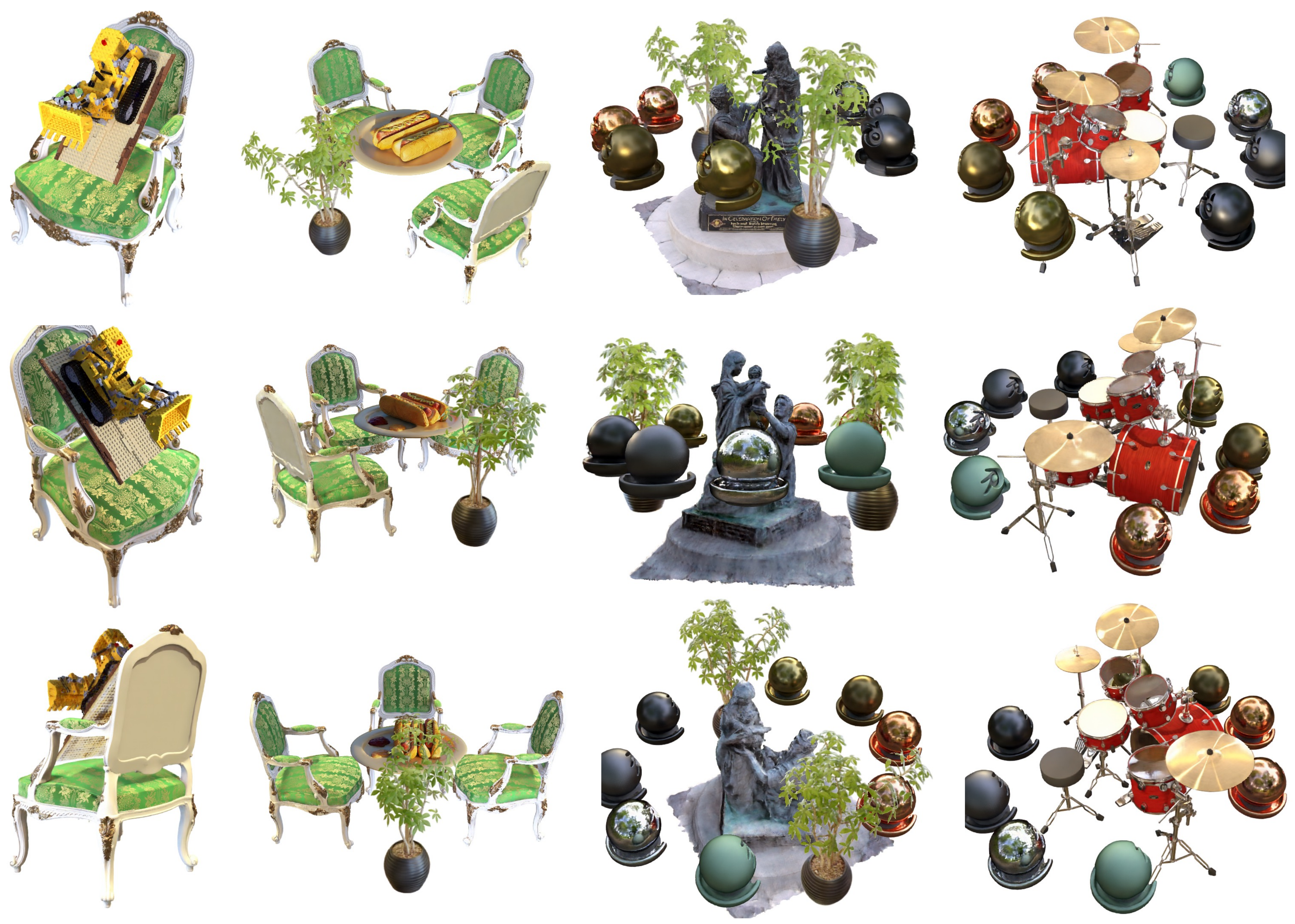}
    \vspace{-0.1in}
    \caption{Scene compostion on NeRF-Synthetic~\cite{mildenhall2021nerf} and Tanks and Temples~\cite{knapitsch2017tanks} datasets.
    These scenes only need to be trained separately. And we can edit objects while combining the scenes.}
  \label{fig:edit_ex}
  \vspace{-0.15in}
\end{figure}

We also show the results of editing, including object editing and scene composition, which are shown in Figs.~\ref{fig:edit_toy} and~\ref{fig:edit_ex}, respectively.
As shwon in Fig.~\ref{fig:edit_toy}, we edit objects in complex scenes, including rotation, translation and scaling.
In Fig.~\ref{fig:edit_ex}, our editing algorithm handles complex occlusion relationships. Moreover, object-level editing and scene composition are performed at the same time in the last two examples in Fig.~\ref{fig:edit_ex}, \emph{i.e.}, the balls in Materials scene are moved and composed with other scenes.
More editing results can be found in supplementary material.

\section{Conclusion}
We develop a novel point cloud rendering pipeline which enables high fidelity reconstruction, real-time rendering and user-friendly editing.
For the novel view synthesis task, experiments on major benchmarks demonstrate the proposed method outperforms existing point cloud rendering methods and achieves the state-of-the-art.
\vspace{-0.15in}
\paragraph{Limitations and Future Work.}
Our current editing method cannot relight the edited objects (\emph{i.e.}, the transformed objects retain their original appearance), handle non-rigid object and appearance editting.
Moreover, the rendering speed of the edited scenes will be reduced.
We will improve these shortcomings in the future.
\vspace{-0.15in}
\paragraph{Acknowledgment}
This work was supported by National Science Foundation of China (U20B2072, 61976137).
This work was also partly supported by SJTU Medical Engineering Cross Research Grant YG2021ZD18.

%%%%%%%%% REFERENCES
{\small
\bibliographystyle{ieee_fullname}
\bibliography{egbib}
}

\clearpage
\appendix

\section{Implementation Details}
\subsection{End-to-end Rendering Pipeline}
\begin{figure}[!t]
\vspace{-0.1in}
    \centering
    % \fbox{\rule{0pt}{4in} \rule{.9\linewidth}{0pt}}
    \includegraphics[width=\linewidth]{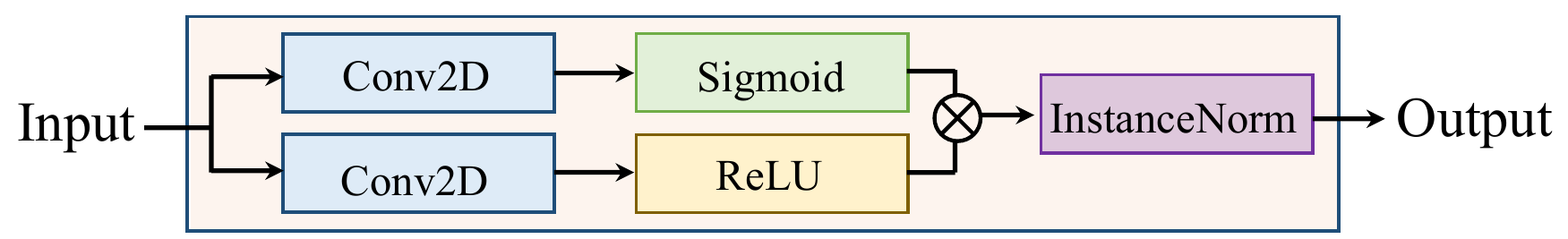}
    \vspace{-0.2in}
    \caption{Gated convolution block.}
  \label{fig:gate}
  \vspace{-0.25in}
\end{figure}

\paragraph{Rasterization Radius.}
During rasterization, each point is expanded to a disk to compensate for point cloud sparsity.  We propose the following two heuristic rules to select disk radius: 1) the disk should cover the gaps between the points, otherwise the
points from the occluded surfaces and the background can be seen through the
front surface (so-called bleeding problem); 2) the radius should be as small as possible, otherwise it will lead to inaccurate depth estimation and the edges of objects will expand.
In practice, the selection of the radius depends on the density of point cloud, and we can adjust the radius by observing the depth map obtained by rasterization.
We also conduct an ablation study on radius, and it can be seen from Tab.~\ref{tab:abla_r} that the effect of the radius is not significant.
\begin{table}[!ht]
\centering
\begin{tabular}{ lccccc } 
 \hline
 radius  & 3e-3 & 4e-3 & 5e-3 & 6e-3 & 7e-3  \\ 
 \hline
 Hotdog  & 34.62 & 35.25 & 35.82 & 35.37 & 34.95  \\ 
 Mic  & 33.64 & 33.49 & 33.40 & 33.14 & 32.73 \\
 \hline
\end{tabular}
\caption{PSNR of Hotdog and Mic scenes under different rasterization radius. The effect of  radius is not significant.}
\label{tab:abla_r}
\end{table}
\vspace{-0.25in}
\paragraph{Radiance Mapping and Refinement.}
We follow the position encoding form of NeRF\cite{mildenhall2021nerf}, but we narrow the encoding interval to provide more fine-grained basis function support for frequency modulation, shown as follows:
% \begin{figure}[!t]
% % \vspace{-0.1in}
%     \centering
%     \includegraphics[width=\linewidth]{img/psnr_hd_mic.png}
%     \caption{PSNR of Mic and Hotdog scenes during training. One Epoch consumes about 12.8s on NeRF-Synthetic. 
% }
%   \label{fig:psnr_hd_mic}
% %   \vspace{-0.1in}
% \end{figure}

\begin{equation}
\begin{aligned}
\gamma(\mathbf{p})=&\left(\sin \left(2^0 \pi \mathbf{p}\right), \cos \left(2^0 \pi \mathbf{p}\right),\right.\\
&\sin \left(2^{0.5} \pi \mathbf{p}\right), \cos \left(2^{0.5} \pi \mathbf{p}\right), \\
& \cdots,  \\
&\left.\sin \left(2^{L-1} \pi \mathbf{p}\right), \cos \left(2^{L-1} \pi \mathbf{p}\right)\right).
\end{aligned}
\end{equation}
We set $L$ as 10 and 2 for spatial coordinates and view directions, respectively.
Our AFNet has five layers, the input feature dimension is 120, and hidden dimensions are 256, 256, 256 and 128. The output dimension is 11 (including 3-channel raw image and 8-channel feature map). Gated convolution block used in U-Net is shown in Fig.~\ref{fig:gate}. For upsampling and downsampling in U-Net, we use nearest neighbor interpolation and average pooling respectively.

\begin{figure*}[!t]
    \centering
    \includegraphics[width=\linewidth]{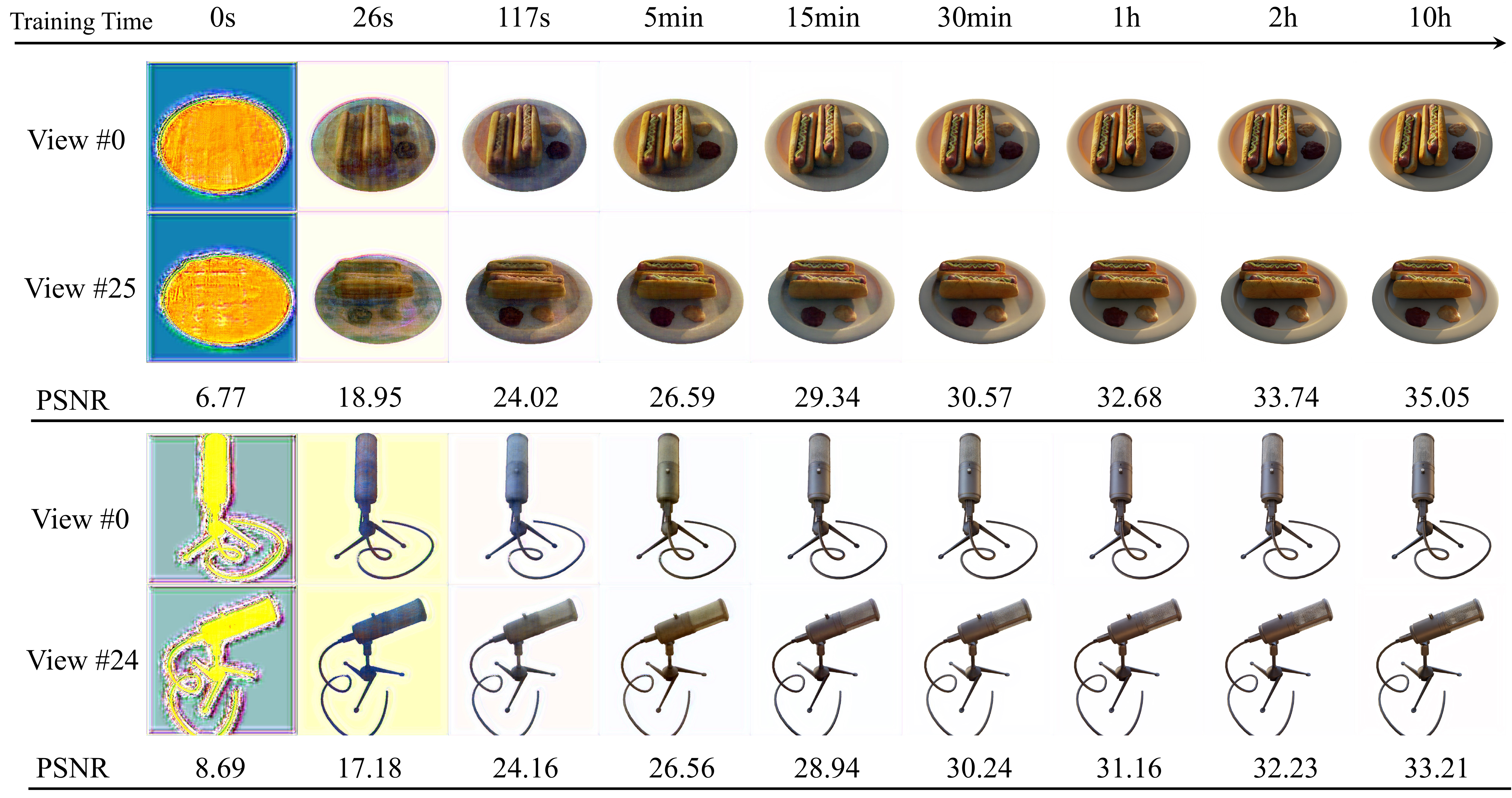}
    \vspace{-0.15in}
    \caption{Rendering results during training. One Epoch consumes about 12.8s on NeRF-Synthetic. 
    Our model requires only ten minutes of training to achieve realistic rendering, but takes about 10 hours to almost converge.
    In order to achieve complete convergence, more than ten hours of fine-tuning is required.
}
  \label{fig:convgence}
\end{figure*}

\vspace{-0.15in}
\paragraph{Training.}
During training stage, we conduct data augmentation by random scaling and
cropping, and we found that random scaling is critical for the training of U-Net.
The loss function is shown as follows:
\begin{equation}
\begin{gathered}
\mathcal{L}=\mathcal{L}_{\text{RGB}} + \lambda \mathcal{L}_{\text{perceptual}}, \\
\end{gathered}
\end{equation}
and we set $\lambda$ as 5e-3 in our experiments.
Thanks to the geometric prior, our model requires only ten minutes of training to achieve realistic rendering, but takes about 10 hours to almost converge, as shown in Fig.~\ref{fig:convgence}.
In order to achieve complete convergence, more than ten hours of fine-tuning is required.

\subsection{Point Cloud Geometry Optimization}
\paragraph{Denoising}
We set $k=16$, \emph{i.e.}, keep 16 points in each pixel buffer. Each point $\mathbf{p}_i$ in point cloud is assigned an opacity parameter $\alpha_i$, and the color $\mathbf{c}_i$ is predicted by MLP.
The ray color is obtained as the following discrete volume rendering equation:
\begin{figure}[!t]
    \vspace{-0.15in}
    \centering
    % \fbox{\rule{0pt}{4in} \rule{.9\linewidth}{0pt}}
    \includegraphics[width=\linewidth]{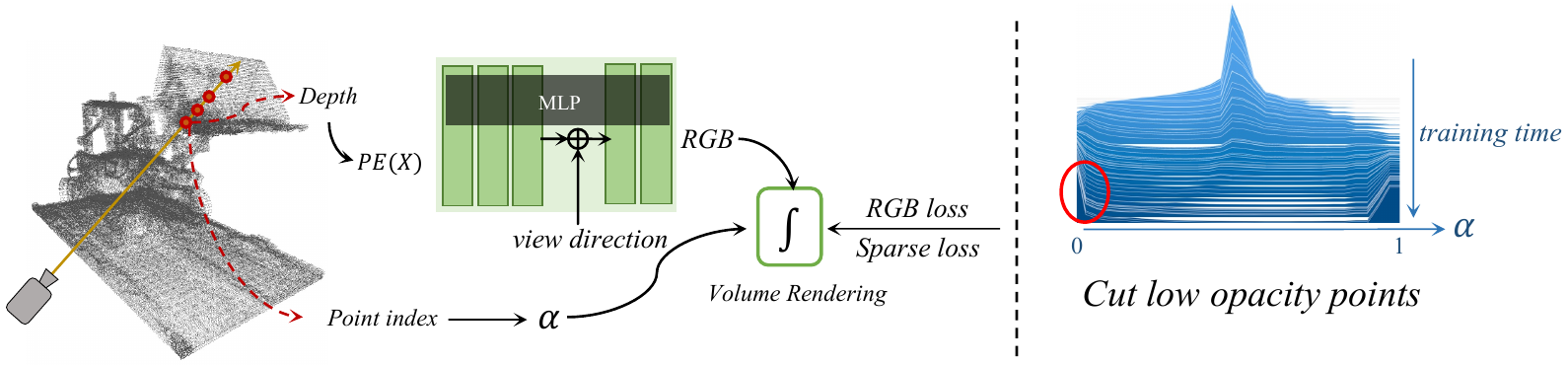}
    \vspace{-0.15in}
    \caption{Denoising step of preprocessing.}
  \label{fig:pcopt}
  % \vspace{-0.15in}
\end{figure}
\begin{figure}[!t]
    \centering
    % \fbox{\rule{0pt}{2in} \rule{.9\linewidth}{0pt}}
    \includegraphics[width=\linewidth]{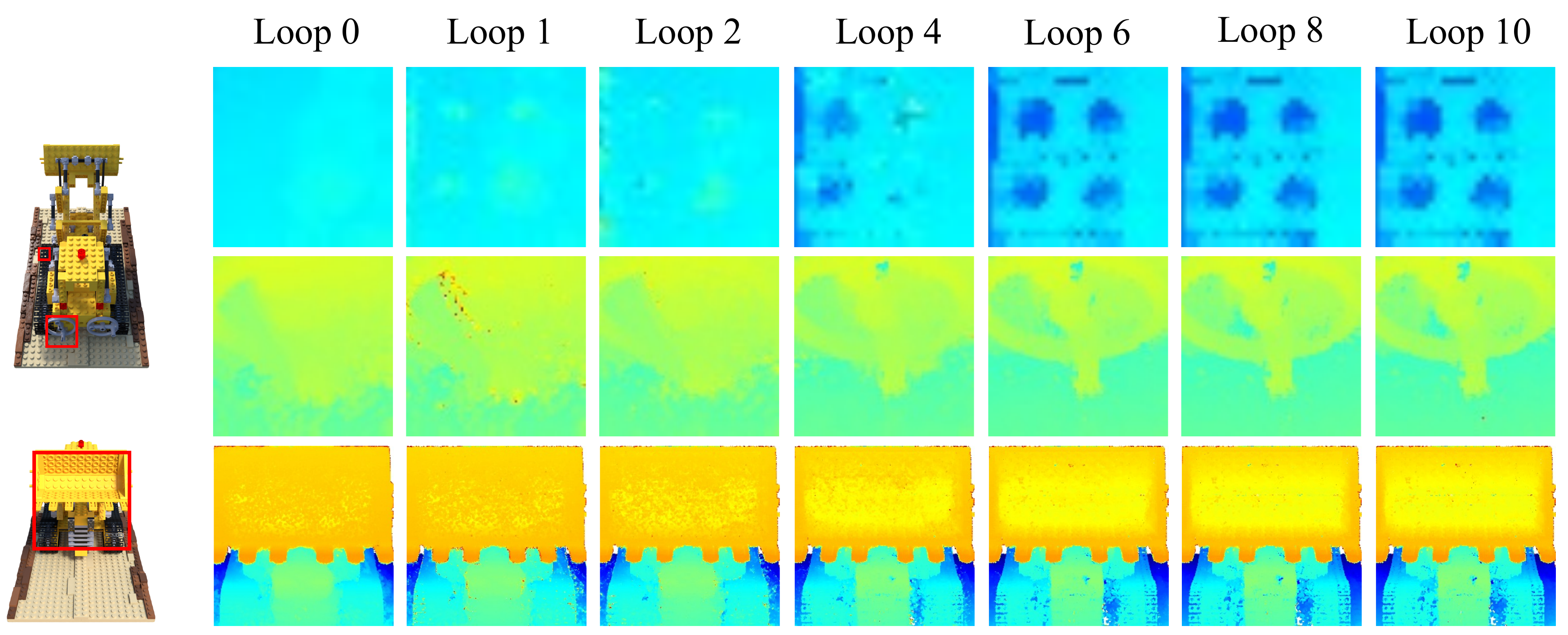}
    \caption{Depth map during geometry optimization. We remove outliers near the track and shovel.}
    \vspace{-0.2in}
  \label{fig:pc_denoi}
\end{figure}
\begin{figure}[t]
    \centering
    % \fbox{\rule{0pt}{2in} \rule{.9\linewidth}{0pt}}
    \includegraphics[width=\linewidth]{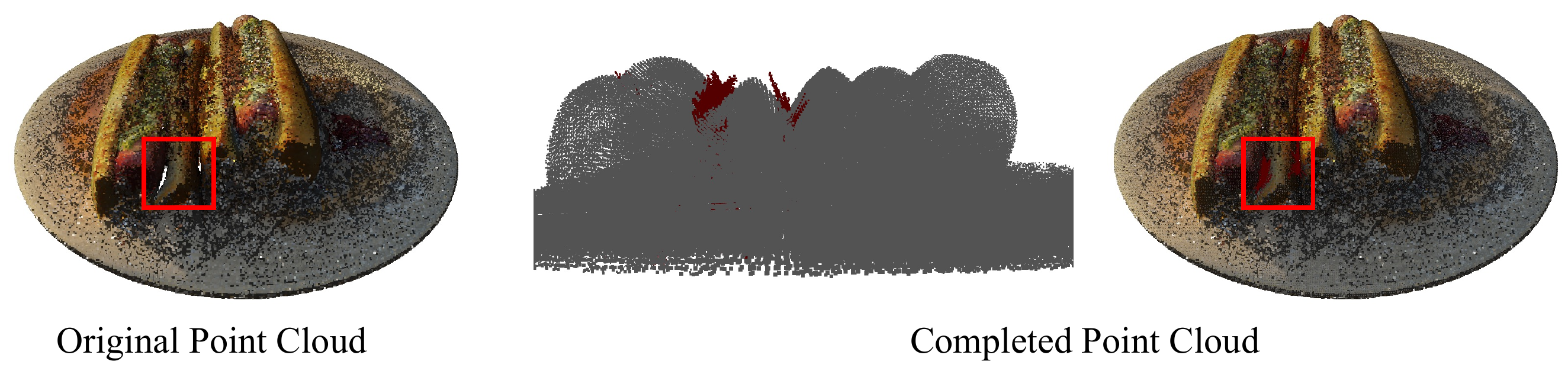}
    \caption{Original and completed point cloud. When an empty buffer is detected, we add a set of points with opacity parameters along the pixel ray in the point clouds, shown as red points. Although the completion step introduces additional noise, it would be removed in the denoising stage of the next iteration.}
  \label{fig:pc_complete}
  \vspace{-0.2in}
\end{figure}
\begin{equation}
\hat{C}(\mathbf{r})=\sum_{i=0}^{k-1} T_i\alpha_i \mathbf{c}_i,\quad
% \text { where } \quad T_i=\exp \left(-\sum_{j=0}^{i-1} \sigma_j \delta_j\right),
\text{where}~~ T_i=\prod_{j=0}^{i-1}(1-\alpha_j).
\end{equation}
We optimize the L2 distance of predicted image and ground truth with sparse regularization as follows:
\begin{equation}
\begin{gathered}
\mathcal{L}=\mathcal{L}_{\text{RGB}} + \lambda_{\text{sparse}} \mathcal{L}_{\text{sparse}}, \\
\mathcal{L}_{\text{RGB}} = \big|\big|\hat{C} - C_{gt} \big|\big| _2 ^2, \\
\mathcal{L}_{\text {sparse }}=\frac{1}{N} \sum_{i=1}^N\left[\log \left(\alpha_i\right)+\log \left(1-\alpha_i\right)\right].
\end{gathered}
\end{equation}
We set $\lambda_{\text{sparse}} = \text{5e-4}$ in our experiments, and the learning rate of MLP and opacity parameters $\alpha$ are set as 5e-4 and 0.01, respectively.
And for scenes without background, we also add transparency loss, \emph{i.e.} L2 distance of predicted transparency and ground truth transparency.
When the training converges, we remove those low-opacity
points, as shwon in Fig. \ref{fig:pcopt}.

\begin{figure*}[!t]
    \centering
    % \fbox{\rule{0pt}{4in} \rule{.9\linewidth}{0pt}}
    \includegraphics[width=\linewidth]{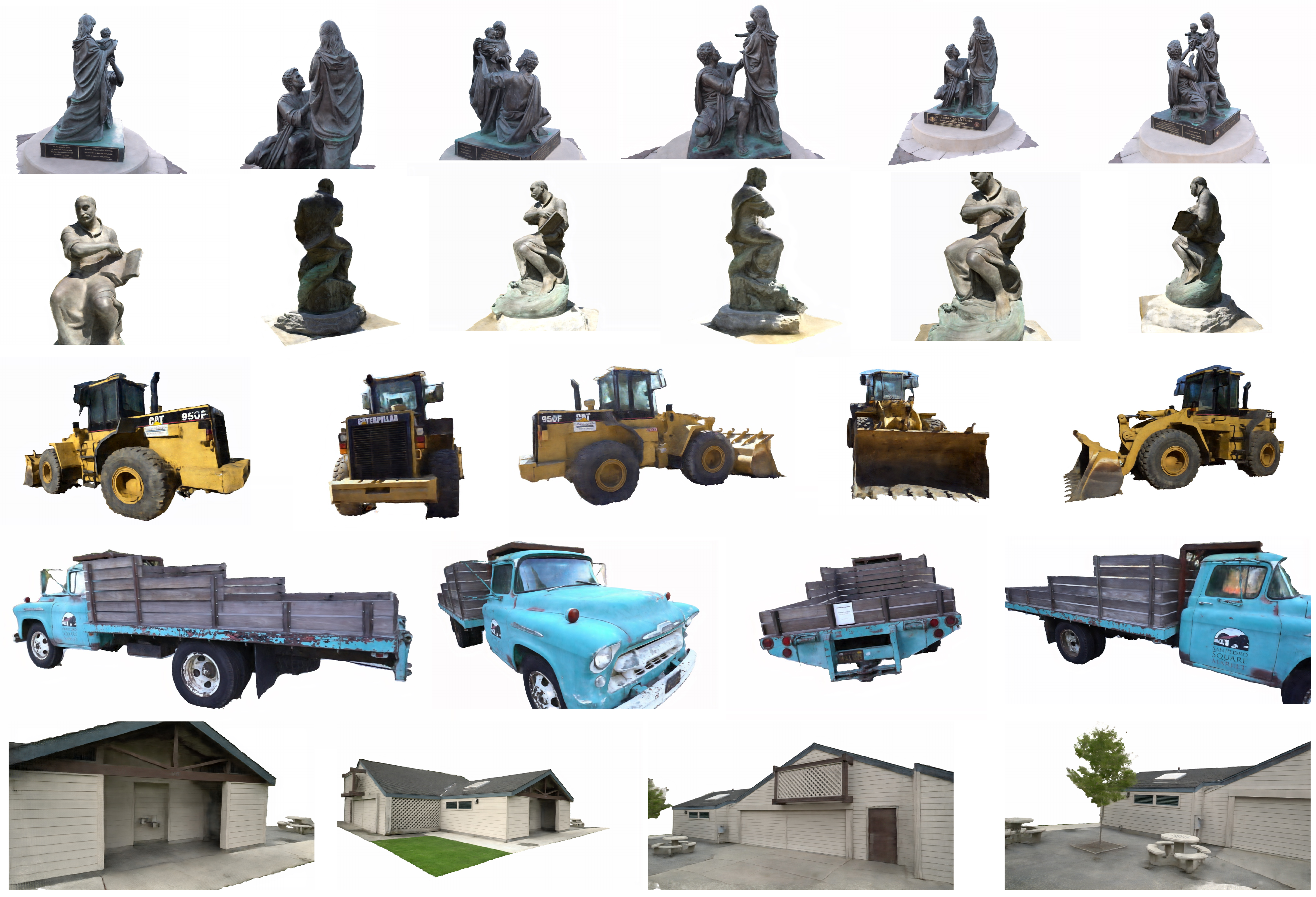}
    \caption{Some rendering results on test views of Tanks and Temples dataset.}
  \label{fig:benchmark_tt}
\end{figure*}

In experiments, we perform 4-20 loops (depends on the raw point cloud quality) of point cloud denoising and completion step alternately. For every denosing step, we train for 300 Epochs.
Fig.~\ref{fig:pc_denoi} shows the depth map computed by volume rendering equation (replace colors as depth values of sample points) during optimization.
Fig.~\ref{fig:pc_complete}  shows the point cloud after completion. Although the completion step introduces additional noise, it would be removed in the denoising stage of the next iteration.
\subsection{Editing Details}
\begin{figure}[!t]
    \centering
    % \fbox{\rule{0pt}{2in} \rule{.9\linewidth}{0pt}}
    \includegraphics[width=\linewidth]{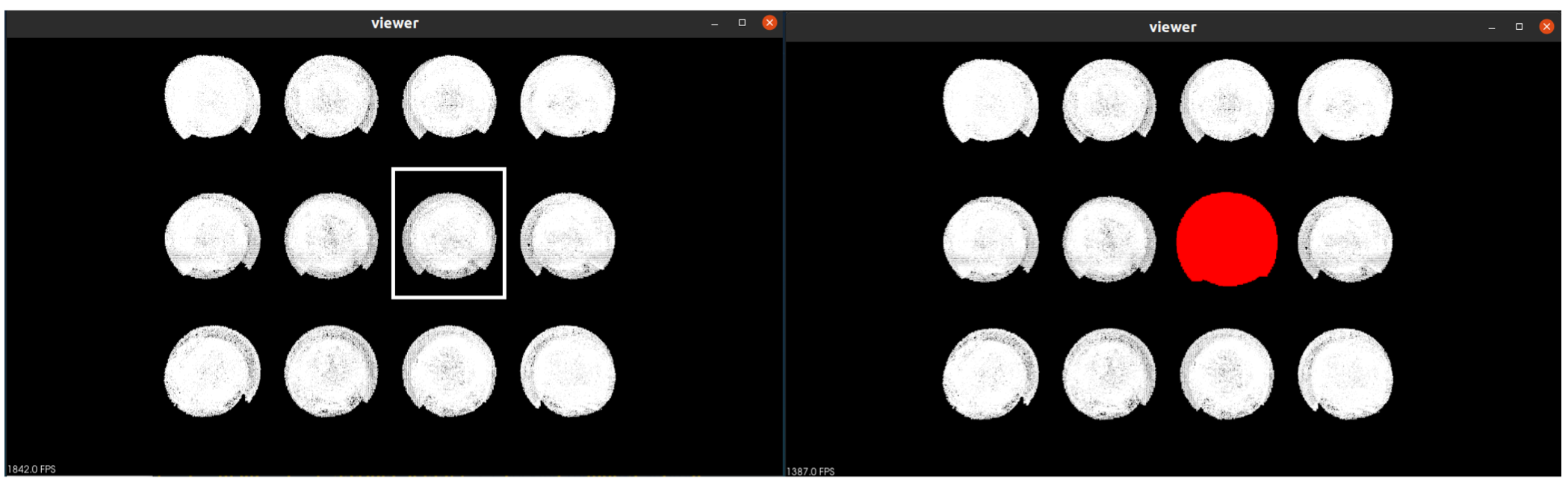}
    \caption{Point cloud selection interface.}
  \label{fig:select}
\end{figure}

In practice, the part of point cloud that needs to be edited (\emph{i.e.}, the object) is stored in the form of a mask. We use PCL library~\cite{Rusu_ICRA2011_PCL} to implement a simple interactive selection function, as shown in Fig.~\ref{fig:select}.
In fact, users can obtain this mask by selecting points using any interactive software.

For scene composition, since the points on the edge of the object will expand, the mask of the scene will also expand outward. We design a simple strategy to avoid artifacts at edges by shrinking the masks determined by the index buffers, please refer to the code for specific implementation.

\section{Experimental Details}
\subsection{Datasets}

\begin{itemize}
\item {NeRF-Synthetic}\cite{mildenhall2021nerf} is a
high quality synthetic dataset containing pathtraced images
of 8 objects.
For each object, there are 100 frames for training and 200 frames for testing. 
The initial point clouds we used are generated by MVSNet\cite{yao2018mvsnet}. 
We set the training size as 800$\times$800 and 
the scaling factor is $[0.5, 1.5]$ of the side length.  At test time, we render with the original resolution 800$\times$800.
\item We use a subset of {Tanks and Temples}\cite{knapitsch2017tanks} dataset, which is from NSVF\cite{liu2020neural}, containing five scenes of real objects. We also use the foreground masks provided by NSVF. Each scene contains 152-384 images of size 1920$\times$1080.
Due to the size limitation of U-Net, we resize the test resolution to 1920$\times$1056, and training size is 640$\times$640. The initial point clouds we used are provided by MVSNet\cite{yao2018mvsnet}. 

\item {ScanNet}\cite{dai2017scannet} is a RGBD dataset of indoor scenes. We evaluate on  $scene0000\_00,$~ $scene0043\_00,$ and $scene0045\_00$, as done in NPBG++\cite{rakhimov2022npbg++} and follow their training-test split.
Specifically, if there are more than 2000 frames in the
scene, we select every 20-th image such that training views would not be too
sparse. Otherwise, we take 100 frames with an equal interval in the image
stream.  Then, we select 10 frames at a fixed interval for testing and the rest for training. We set the training size as 720$\times$720 and the test resolution is
960$\times$1200. The initial point clouds are obtained by the provided depth maps.  

\item {DTU}\cite{jensen2014large} is a multi-view stereo dataset with a resolution of 1200$\times$1600. 
We evaluate on $scan110$, $scan114$ and $scan118$, and use the same point clouds and training-test split as NPBG++\cite{rakhimov2022npbg++}. We mask out the background in training using the binary segmentation masks provided by IDR\cite{yariv2020multiview}. The training patch size is set as 800$\times$800 and
the test resolution 1200$\times$1600.

\item For ToyDesk~\cite{yang2021learning} dataset,  we only perform training and editing without evaluation.

\end{itemize}

\begin{figure}[!t]
    \centering
    \includegraphics[width=\linewidth]{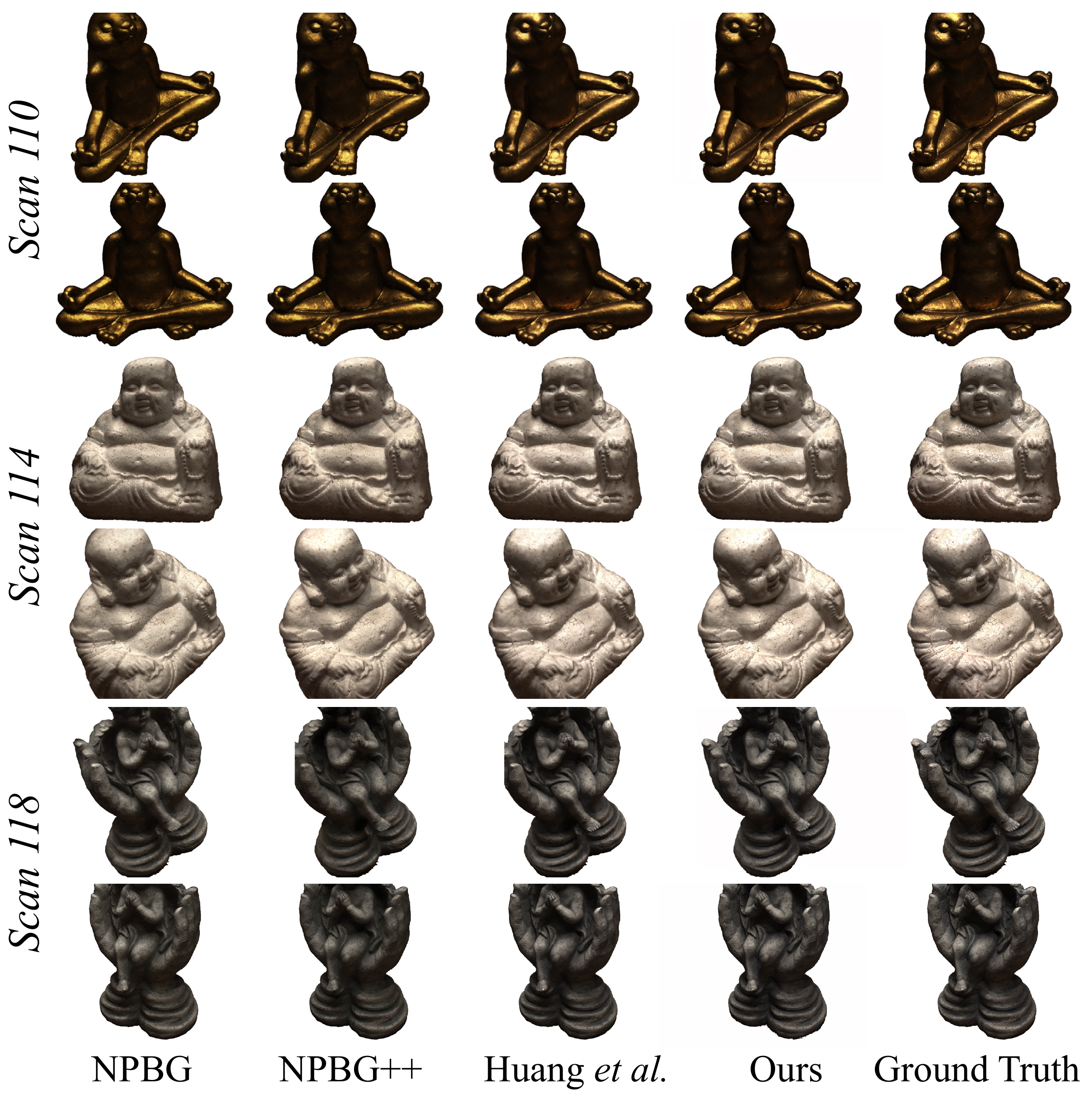}
    \caption{Qualitative comparison on DTU dataset.}
    \label{fig:dtu}
\end{figure}
\begin{figure}[!t]
    \centering
    \includegraphics[width=\linewidth]{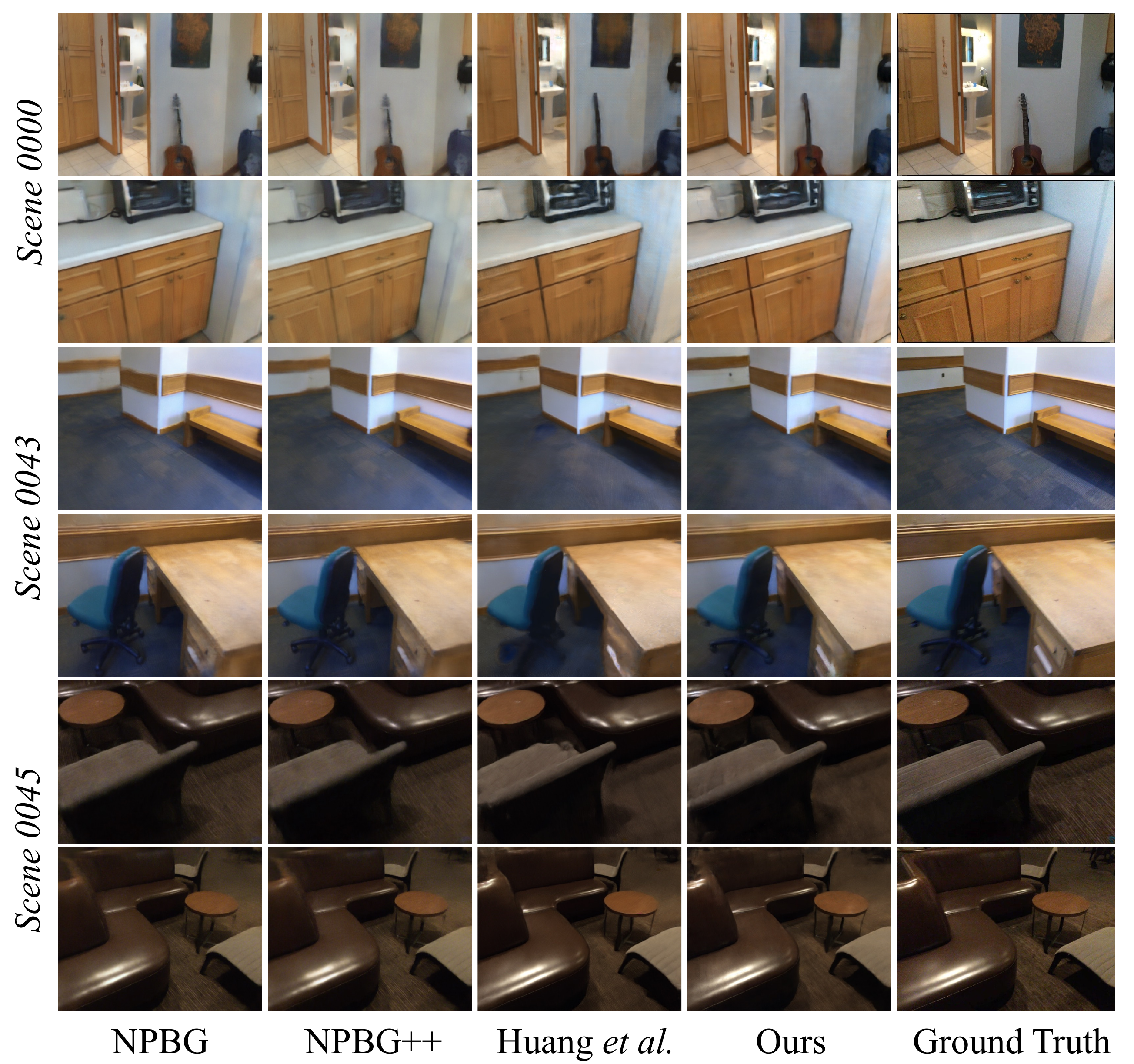}
    \caption{Qualitative comparison on ScanNet dataset.}
    \label{fig:scan}
\end{figure}
\begin{figure}[!t]
    \centering
    \includegraphics[width=\linewidth]{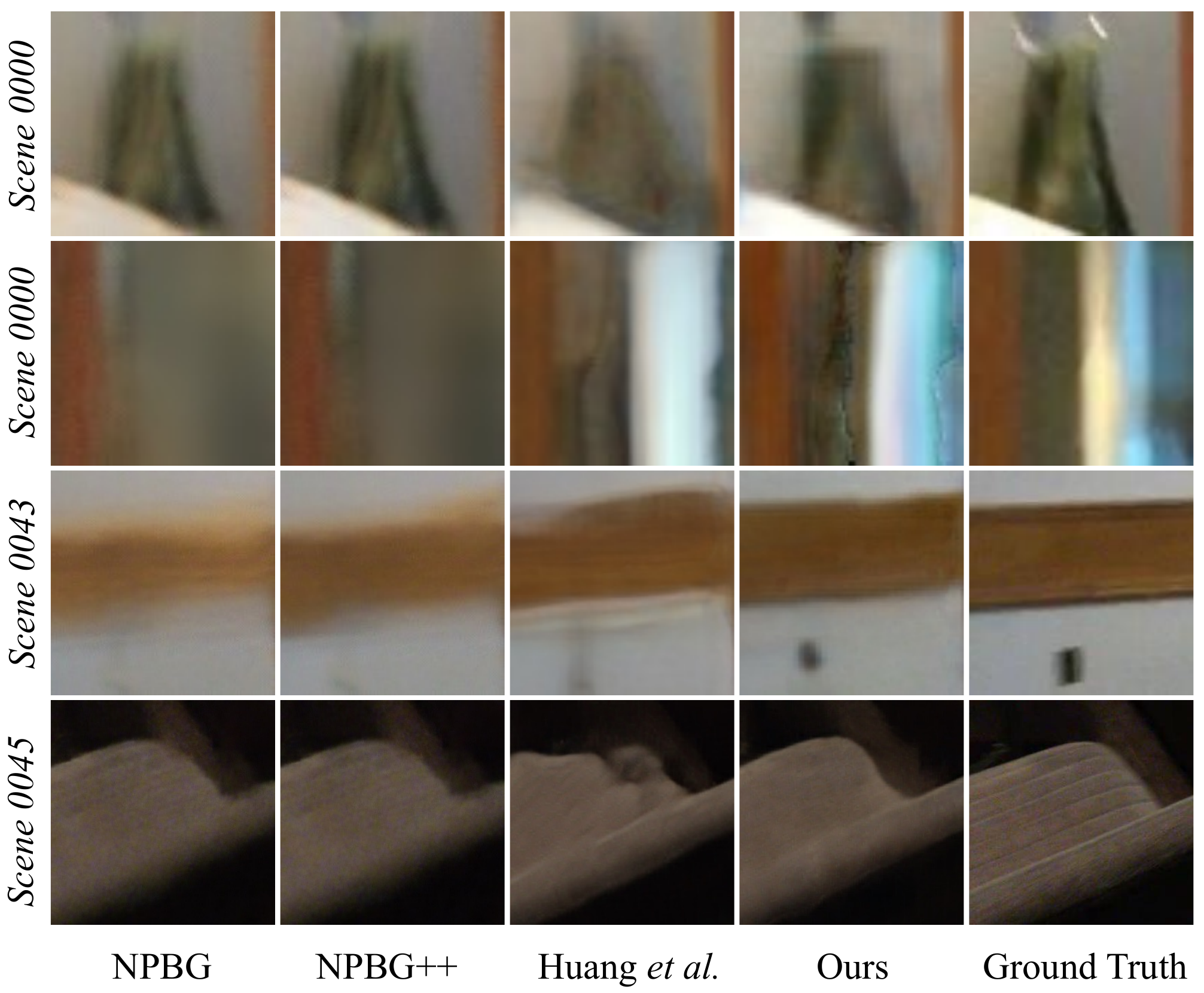}
    \caption{Qualitative comparison in detail on ScanNet dataset. Our rendering results are closer to ground truth.}
    \label{fig:scan_detail}
    \vspace{-0.15in}
\end{figure}
\vspace{-0.15in}
\subsection{Compared Methods}
\begin{itemize}
  \item {NPBG}\cite{aliev2020neural}: A famous point-based rendering method, which uses point-wise features to encode the appearance of each surface point and an U-Net for decoding.
  \item {NPBG++}\cite{rakhimov2022npbg++}: The improved version of NPBG, which predicts the descriptors with a feature extractor and makes the neural descriptors view-dependent.
  \item {Huang \textit{et al}.}\cite{oursaaai}: The state-of-the-art point-based neural rendering, which combines explicit point clouds and implicit radiance mapping. However, its performance is still lower than that of NeRF, due to the weak frequency expressiveness and lack of geometric optimization.
  \item {CCNeRF}\cite{tang2022compressible}: The latest editable variant of NeRF\cite{mildenhall2021nerf}, which represents a 3D scene as a compressible 3D Tensor. Due to the massive sampling and calculation of tensor decomposition, the rendering speed is only 1.05 FPS. 
\end{itemize}
\subsection{Detailed Results}
\begin{figure}[!t]
    \centering
    % \fbox{\rule{0pt}{2in} \rule{.9\linewidth}{0pt}}
    \includegraphics[width=\linewidth]{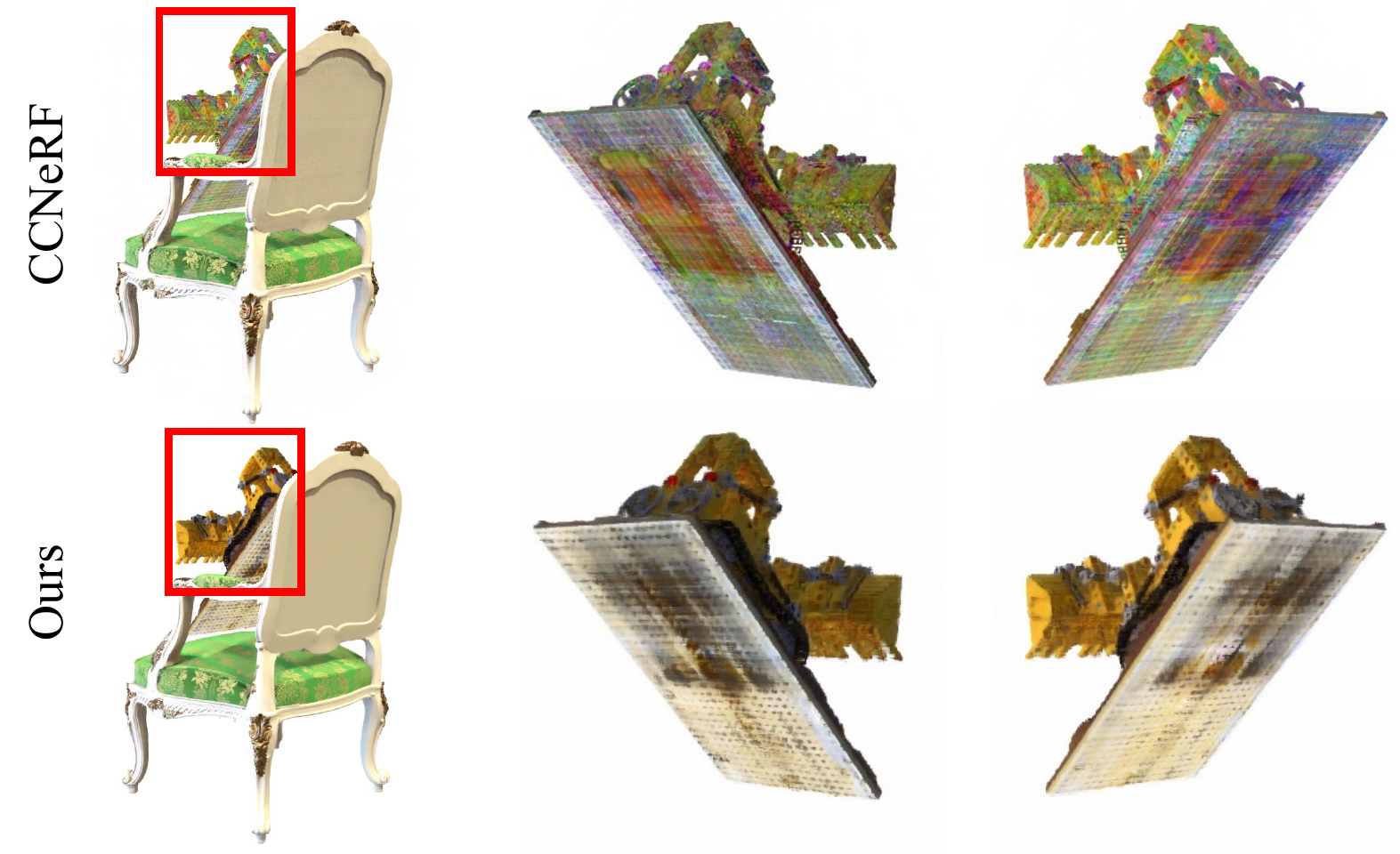}
    \vspace{-0.2in}
    \caption{Extreme views of Lego scene that may be encountered in editing. We render reasonable results while CCNeRF renders distorted colors.}
  \label{fig:exview_lego}
  \vspace{-0.2in}
\end{figure}

\begin{figure*}[!htbp]
    \centering
    % \fbox{\rule{0pt}{8in} \rule{.9\linewidth}{0pt}}
    \includegraphics[width=\linewidth]{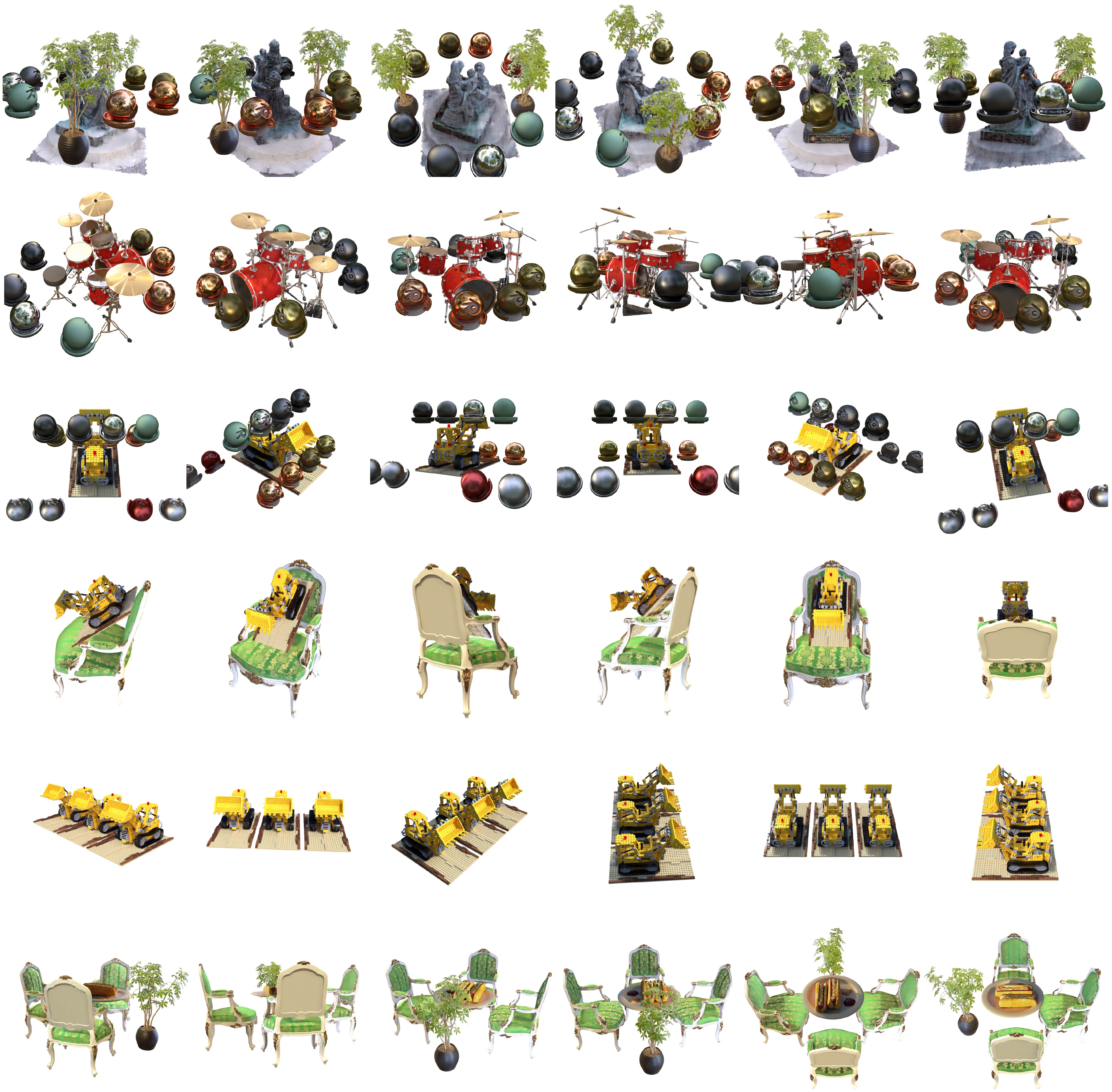}
    \caption{Editing results, including object-level editing and scene composition.}
  \label{fig:edit_nerf}
\end{figure*}

\begin{figure*}[!htbp]
    \centering
    % \fbox{\rule{0pt}{8in} \rule{.9\linewidth}{0pt}}
    \includegraphics[width=\linewidth]{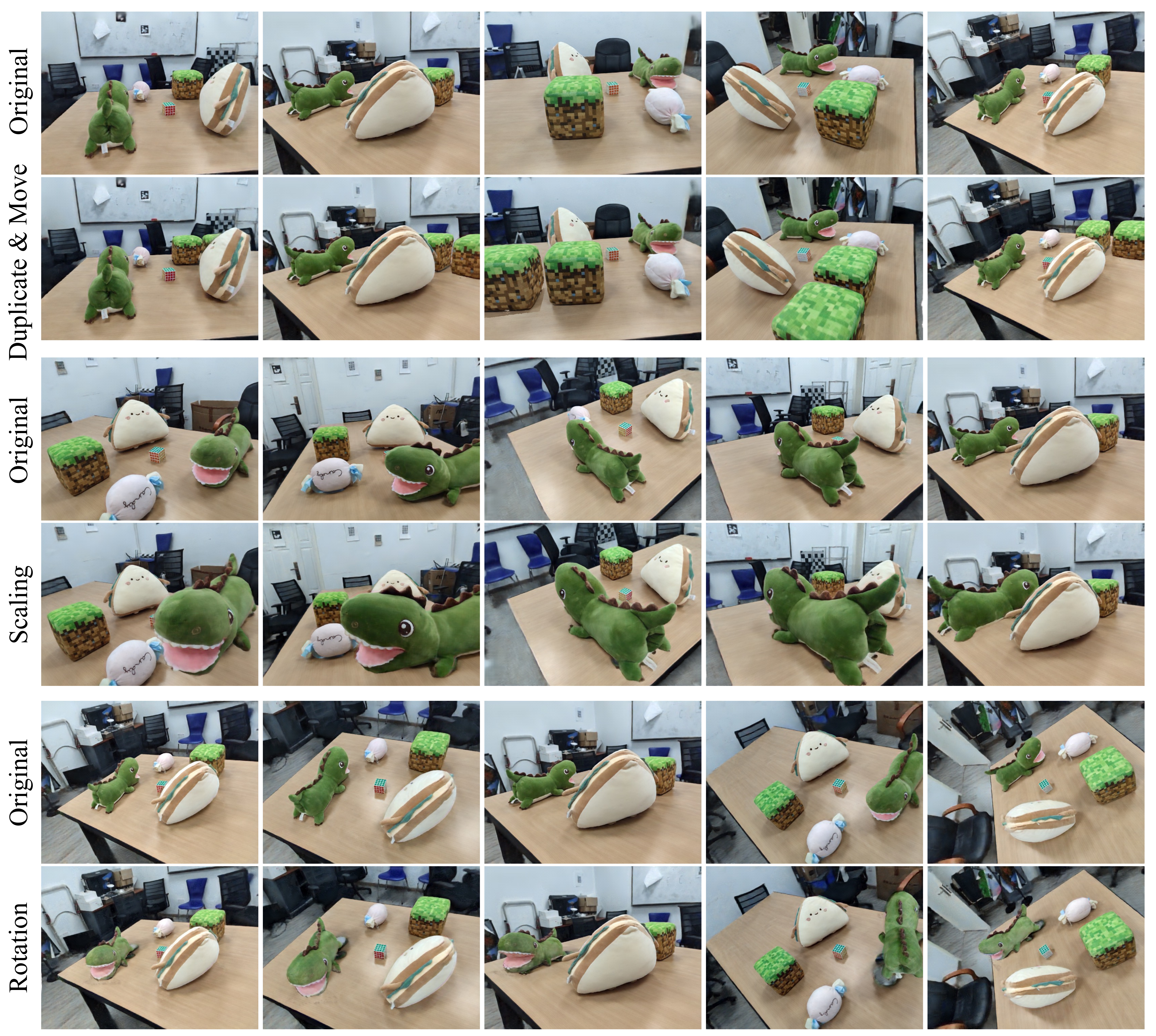}
    \caption{Object translation, rotation and scaling on ToyDesk\cite{yang2021learning} dataset. As can be seen from the last line of results, the rotation of the green object will expose the untrained local space, resulting in artifacts, which are also reflected in the results of Object-NeRF\cite{yang2021learning}.}
  \label{fig:edit_toy_supp}
\end{figure*}

% \paragraph{Quantitative results.}
We present some rendering results on Tanks and Temples dataset in Fig.~\ref{fig:benchmark_tt}, and qualitative comparisons on DTU and ScanNet datasets in Figs.~\ref{fig:dtu} and \ref{fig:scan}, respectively.
A comparison of details on ScanNet dataset is shwon in Fig. \ref{fig:scan_detail}. 
Although the rendering results look similar overall as seen from Figs.~\ref{fig:dtu} and \ref{fig:scan}, we are better at some details, as shwon in Fig.~\ref{fig:scan_detail}.
Due to the blurring of some training images in ScanNet dataset, some rendering results of novel views are also blurred.  For the compared methods, we all use the same experimental configuration, which is fair.
% \paragraph{Qualitative results.}
We present quantitative evaluation for each scene of each dataset in Tabs. \ref{tab:benchmark_nerf}, \ref{tab:benchmark_tt} and \ref{tab:benchmark_dtu_scan}.
More editing results are shown in Figs. \ref{fig:exview_lego}, \ref{fig:edit_nerf} and \ref{fig:edit_toy_supp}.


\begin{table*}[!ht]
    \centering
    \small
    \setlength\tabcolsep{1pt}
    % \footnotesize
    \vspace{-8pt}
    \begin{tabular*}{\hsize}{@{}@{\extracolsep{\fill}}lccccccccc@{}}
    \toprule \multicolumn{10}{c}{NeRF-Synthetic} \\
     Method & Chair & Drums & Ficus & Hotdog 
    & Lego & Materials & Mic & Ship & Mean  \\
    \midrule
    \multicolumn{10}{c}{PSNR$\uparrow$} \\
    \midrule
    NeRF\cite{mildenhall2021nerf} & 33.00 & 25.01 & 30.13 & \textbf{36.18} & 32.54 & 29.62 & 32.91 & \textbf{28.65} & 31.01\\
    CCNeRF-CP\cite{tang2022compressible}  & 33.63 &	24.23 &	29.40 &	35.27 &	32.94 &	28.34 &	32.81 &	27.77 &	30.55   \\
    CCNeRF-HY-S\cite{tang2022compressible}& \textbf{34.37} & 	24.76 & 	30.04 & 	36.04& 	\textbf{33.66}& 	28.96 &	{33.53} &	28.38 &	31.22 \\
    NPBG\cite{aliev2020neural} 
    & 28.81 & 23.57 & 28.23 & 32.03 & 27.72 & 27.24 & 31.16 & 26.04 & 28.10\\
    NPBG++\cite{rakhimov2022npbg++} 
    & 28.72 & 23.60 & 28.11 & 32.22 & {27.84} & 27.12 & 31.23 & 26.11 & 28.12\\
    Huang \textit{et al.} 
    & {31.13} & {24.51} & {29.09} &  {33.20} & 26.62 &  {28.03} &  {32.94} &  {26.14} &  {28.96}\\
   
    Ours & 33.06 & \textbf{25.95} & \textbf{32.19} & 35.82 & 31.56 & \textbf{29.69} & \textbf{33.64} & 27.97  & \textbf{31.24}\\

    \midrule
    \multicolumn{10}{c}{SSIM$\uparrow$} \\
    \midrule
    NeRF\cite{mildenhall2021nerf} 
    & 0.967 & 0.925 & 0.964 & 0.974 &  0.961 & 0.949 & 0.980 & \textbf{0.856} & 0.947\\
    CCNeRF-CP\cite{tang2022compressible} & 0.964 &	0.906 &	0.950 &	0.966 &	0.957 &	0.923 &	0.971 &	0.842 &	0.935 \\
    CCNeRF-HY-S\cite{tang2022compressible} & \textbf{0.976} &	0.918 &	0.962 &	\textbf{0.978} &	\textbf{0.969} &	0.935 	&0.983 &	0.853 &	0.947\\
    
    NPBG\cite{aliev2020neural} 
    & 0.954 & 0.902 & 0.942 & 0.960 & 0.919 & 0.922 & 0.970 & 0.812 & 0.923   \\
    NPBG++\cite{rakhimov2022npbg++} 
    &  {0.961} & 0.910 & 0.947 & 0.961 &  {0.923} & 0.925 & 0.972 & 0.822 & 0.928  \\
    Huang \textit{et al.}  
    & 0.953 &  {0.924} &  {0.958} &  {0.964} & 0.902 &  {0.945} &  {0.983} &  {0.824} &  {0.932}  \\
    
    Ours & 0.974 &	\textbf{0.938} &	\textbf{0.971} &	0.974 &	0.956 &	\textbf{0.955} &	\textbf{0.986} &	0.845 & \textbf{0.950}  \\
    \midrule
    \multicolumn{10}{c}{LPIPS$\downarrow$} \\
    \midrule
    NeRF\cite{mildenhall2021nerf} 
    & 0.046 & 0.091 & 0.044 & 0.121 
    & 0.050 & 0.063 & 0.028 & 0.206 & 0.081\\
    CCNeRF-CP\cite{tang2022compressible}&0.037 &	0.111 &	0.055 &	0.057 &	0.037 &	0.082 &	0.031 &	0.196 &	0.076  \\
    CCNeRF-HY-S\cite{tang2022compressible} & 0.036 &	0.109 	&0.054 	&0.056 &	\textbf{0.036} &	0.080 &	0.030 &	0.192 &	0.074 \\
    NPBG \cite{aliev2020neural} 
    & 0.047 & 0.093 & 0.046 & 0.055 & 0.089 & 0.076 & 0.037 & 0.171 & 0.077    \\
    NPBG++\cite{rakhimov2022npbg++} 
    & 0.048 & 0.092 & 0.043 & 0.053 & 0.089 & 0.074 & 0.030 & 0.178 & 0.076 \\
    Huang \textit{et al.}  
    &  {0.040} &  {0.068} &  {0.035} &  {0.038} &  {0.085} &  {0.050} &  \textbf{0.014} &  {0.159} & {0.061} \\
    Ours & \textbf{0.025} & \textbf{0.065} & \textbf{0.026} & \textbf{0.028} & 0.045 & \textbf{0.046} & {0.015} & \textbf{0.142} & \textbf{0.049} \\

    \bottomrule

    \end{tabular*}
    \caption{PSNR$\uparrow$, SSIM$\uparrow$ and LPIPS$\downarrow$ on each scene of NeRF-Synthetic dataset.}
    \label{tab:benchmark_nerf}
    % \vspace{-10pt}
\end{table*}

\begin{table*}[!ht]
    \small
    \centering
    \setlength\tabcolsep{1pt}
    % \footnotesize
    \vspace{-8pt}
    \begin{tabular*}{\hsize}{@{}@{\extracolsep{\fill}}lcccccc@{}}
    \toprule \multicolumn{7}{c}{Tanks and Temples} \\
     Method & Barn & Caterpillar & Family & Ignatius & Truck & Mean  \\
    \midrule
    \multicolumn{7}{c}{PSNR$\uparrow$} \\
    \midrule
    NeRF\cite{mildenhall2021nerf} & 24.05 & 23.75 & 30.29 & 25.43 & 25.36 & 25.78 \\
    % NSVF &  27.16 & 26.44 & 33.58 & 27.91 & 26.92 & 28.40 \\
    CCNeRF-CP\cite{tang2022compressible} & 25.84 &	24.02 &	32.13 &	27.24 &	25.84 &	27.01  \\
    CCNeRF-HY-S\cite{tang2022compressible} & 26.34 & 24.48 &	\textbf{32.75} &	27.76 &	26.34 &	27.53 \\
    NPBG\cite{aliev2020neural} & 24.86 &	22.05 &	30.84 &	26.50 &	25.59 &	25.97 \\
    NPBG++\cite{rakhimov2022npbg++} & 24.90 &	22.22 &	30.67 &	26.98 &	25.45 &	26.04\\
    Huang \textit{et al.}\cite{oursaaai} & 25.34&	23.09&	30.65&	27.01&	25.68&	26.35 \\
    Ours & \textbf{27.01} &	\textbf{24.67} &	32.36 &	\textbf{28.83} &	\textbf{26.56} &	\textbf{27.79} \\
    \midrule
    \multicolumn{7}{c}{SSIM$\uparrow$} \\
    \midrule 
    NeRF\cite{mildenhall2021nerf} & 0.750 & 0.860 & 0.932 & 0.920 & 0.860 & 0.864 \\
    % NSVF & 0.823 & 0.900 & 0.954 & 0.930 & 0.895 & 0.900\\
    CCNeRF-CP\cite{tang2022compressible} & 0.807 &	0.864 &	0.934 &	0.916 &	0.872 &	0.879  \\
    CCNeRF-HY-S\cite{tang2022compressible} & 0.827 & \textbf{0.886} &	\textbf{0.957} &	0.939 &	0.894 &	0.901  \\
    NPBG\cite{aliev2020neural} & 0.841 &	0.848 &	0.940 &	0.928 &	0.887 &	0.889  \\
    NPBG++\cite{rakhimov2022npbg++} & 0.842 &	0.863 &	0.943 &	0.933 &	0.878 &	0.892  \\
    Huang \textit{et al.}\cite{oursaaai} & 0.841 &	0.855 &	0.946 &	0.936 &	0.886 &	0.893  \\
    Ours & \textbf{0.847} &	0.876 &	0.953 &	\textbf{0.941} &	\textbf{0.895} &	\textbf{0.902} \\
    \midrule
    \multicolumn{7}{c}{LPIPS$\downarrow$} \\
    \midrule 
    NeRF\cite{mildenhall2021nerf} & 0.395 & 0.196 & 0.098 & 0.111 & 0.192 & 0.198\\
    % NSVF & 0.307 & 0.141 & 0.063 & 0.106 & 0.148 & 0.153 \\
    CCNeRF-CP\cite{tang2022compressible}& 0.310 &	0.223 &	0.078 &	0.099 &	0.192 &	0.180 \\
    CCNeRF-HY-S\cite{tang2022compressible} &0.304&	0.219&	0.076&	0.097&	0.188&	0.177 \\
    NPBG\cite{aliev2020neural} & 0.219 & 	0.182 &	0.072 &	0.076 &	0.138 &	0.137 \\
    NPBG++\cite{rakhimov2022npbg++}& 0.197  &	0.181 &	0.075 &	0.068 &	0.131 &	0.130 \\
    Huang \textit{et al.}\cite{oursaaai}& 0.195&	0.179&	0.069&	0.077&	0.131&	0.130 \\
     
    Ours & \textbf{0.179} &	\textbf{0.175} &	\textbf{0.066} &	\textbf{0.073} &	\textbf{0.131} &	\textbf{0.125} \\

    \bottomrule
    \end{tabular*}
    \caption{PSNR$\uparrow$, SSIM$\uparrow$ and LPIPS$\downarrow$ on each scene of Tanks and Temples dataset.}
    \label{tab:benchmark_tt}
    % \vspace{-10pt}
\end{table*}

\begin{table*}[!ht]
    \centering
    \setlength\tabcolsep{1pt}
    % \footnotesize
    \vspace{-8pt}
    \begin{tabular*}{\hsize}{@{}@{\extracolsep{\fill}}lcccccccc@{}}
    \toprule & \multicolumn{4}{c}{ScanNet} & \multicolumn{4}{c}{DTU} \\
     Method & Scene0000 & Scene0043 & Scene0045 & Mean & Scan110 & Scan114 & Scan118 & Mean  \\
    \hline
    \multicolumn{9}{c}{PSNR$\uparrow$} \\
    \hline
    NeRF\cite{mildenhall2021nerf} & 22.08 &	25.98 &	29.15 &	25.74 & 25.55 &	27.42 &	\textbf{27.78} &	26.92   \\
    CCNeRF-CP\cite{tang2022compressible}&21.14&	25.89&	26.91&	24.65&25.88&	27.89&	26.61&	26.79  \\
    CCNeRF-HY-S\cite{tang2022compressible}&21.38&	\textbf{26.22}&	27.91&	25.17&26.23&	28.12&	27.34&	27.23 \\
    NPBG\cite{aliev2020neural}&22.24&	25.27&	27.75&	25.09& 24.65&	26.74&	26.62&	26.00 \\
    NPBG++\cite{rakhimov2022npbg++}&22.05&	25.51&	28.26&	25.27&24.84&	26.72&	26.67&	26.08 \\
    Huang \textit{et al.}\cite{oursaaai}&23.79&	25.26&	28.59&	25.88&		25.05&	26.87&	26.75&	26.22 \\
    Ours & \textbf{24.35}&	26.15&	\textbf{29.48}&	\textbf{26.66}&		\textbf{26.45}&	\textbf{28.31}&	27.06&	\textbf{27.27}  \\
    \hline
    \multicolumn{9}{c}{SSIM$\uparrow$} \\
    \hline 
    NeRF\cite{mildenhall2021nerf} & 0.729 &	0.869 &	0.743 &	0.780 	&	0.909 &	0.894 &	0.924 &	0.909  \\
    CCNeRF-CP\cite{tang2022compressible}&0.695&	0.849&	0.779&	0.774&		0.904&	0.894&	0.922&	0.907  \\
    CCNeRF-HY-S\cite{tang2022compressible}&0.701&	0.854&	0.788&	0.781&		0.909&	\textbf{0.896}&	\textbf{0.925}&	\textbf{0.910}  \\
    NPBG\cite{aliev2020neural}&0.695&	0.830&	0.686&	0.737&		0.907&	0.873&	0.904&	0.895  \\
    NPBG++\cite{rakhimov2022npbg++}&0.742&	0.859&	0.716&	0.772&		0.910&	0.865&	0.909&	0.895 \\
    Huang \textit{et al.}\cite{oursaaai}&0.748&	0.844&	0.789&	0.794&		\textbf{0.917}&	0.863&	0.919&	0.900 \\
    Ours & \textbf{0.754}&	\textbf{0.863}&	\textbf{0.792}&	\textbf{0.803}&		0.915&	0.892&	0.916&	0.908 \\
    \hline
    \multicolumn{9}{c}{LPIPS$\downarrow$} \\
    \hline 
    NeRF\cite{mildenhall2021nerf} &  0.588&	0.466&	0.558&	0.537&		0.194&	0.217&	0.182&	0.198 \\
    CCNeRF-CP\cite{tang2022compressible}&0.599&	0.457&	0.569&	0.542&		0.181&	0.191&	0.161&	0.178 \\
    CCNeRF-HY-S\cite{tang2022compressible}&0.594&	0.456&	0.566&	0.539&		0.171&	0.188&	0.154&	0.171 \\
    NPBG\cite{aliev2020neural}&0.474&	0.421&	0.482&	0.459&		0.124&	0.143&	0.123&	0.130 \\
    NPBG++\cite{rakhimov2022npbg++}&0.457&	0.410&	0.477&	0.448&		0.125&	0.148&	0.121&	0.131 \\
    Huang \textit{et al.}\cite{oursaaai}&0.440&	0.389&	0.415&	0.415&		0.124&	0.154&	\textbf{0.117}&	0.132 \\
    Ours &  \textbf{0.418}&	\textbf{0.369}&	\textbf{0.412}&	\textbf{0.400}&		\textbf{0.122}&	\textbf{0.143}&	0.122&	\textbf{0.129} \\
    \bottomrule
    \end{tabular*}
    \caption{PSNR$\uparrow$, SSIM$\uparrow$ and LPIPS$\downarrow$ on each scene of ScanNet and DTU datasets.}
    \label{tab:benchmark_dtu_scan}
    % \vspace{-10pt}
\end{table*}

%%%%%%%%% REFERENCES
% {\small
% \bibliographystyle{ieee_fullname}
% \bibliography{egbib}
% }

\end{document}